%% file: eccv2022submission.tex
\documentclass[runningheads, oneside, openany]{llncs}
\usepackage{graphicx}

\usepackage{tikz}
\usepackage{comment}
\usepackage{appendix}
\usepackage{amsmath,amssymb} % define this before the line numbering.
\usepackage{color}

\usepackage[accsupp]{axessibility}  % Improves PDF readability for those with disabilities.

\usepackage[width=122mm,left=12mm,paperwidth=146mm,height=193mm,top=12mm,paperheight=217mm]{geometry}

\usepackage{booktabs}
\usepackage{xcolor}
\usepackage[ruled]{algorithm2e}
\usepackage{epsfig}
\usepackage{wrapfig,lipsum}
\usepackage{tabularx}
\usepackage{algpseudocode}
\usepackage{caption}
\usepackage{indentfirst}

\definecolor{citecolor}{HTML}{0071bc}
\usepackage[pagebackref=true,breaklinks=true,colorlinks,citecolor=citecolor,bookmarks=false]{hyperref}

\usepackage[capitalize]{cleveref}
\crefname{section}{Sec.}{Secs.}
\Crefname{section}{Section}{Sections}
\Crefname{table}{Table}{Tables}
\crefname{table}{Tab.}{Tabs.}

\newcommand{\titlecap}[2]{\textbf{#1} #2}

\newcommand{\mypara}[1]{\vspace{1mm}\noindent \textbf{#1}}
\newlength\savewidth\newcommand\shline{\noalign{\global\savewidth\arrayrulewidth
  \global\arrayrulewidth 1pt}\hline\noalign{\global\arrayrulewidth\savewidth}}
\newcommand{\tablestyle}[2]{\setlength{\tabcolsep}{#1}\renewcommand{\arraystretch}{#2}\centering\footnotesize}
\newcommand{\etal}{\textit{et al}.}
\DeclareMathOperator*{\argmin}{arg\,min}

\newcommand\blfootnote[1]{%
  \begingroup
  \renewcommand\thefootnote{}\footnote{#1}%
  \addtocounter{footnote}{-1}%
  \endgroup
}

\begin{document}
% \linenumbers
\pagestyle{headings}
\mainmatter
\def\ECCVSubNumber{5261}  % Insert your submission number here

\title{DexMV: Imitation Learning for Dexterous Manipulation from Human Videos} % Replace with your title

\begin{comment}
\titlerunning{ECCV-22 submission ID \ECCVSubNumber} 
\authorrunning{ECCV-22 submission ID \ECCVSubNumber} 
\author{Anonymous ECCV submission}
\institute{Paper ID \ECCVSubNumber}
\end{comment}
%******************

% \begin{comment}
\titlerunning{DexMV}

\author{Yuzhe Qin\inst{*} \and
Yueh-Hua Wu\inst{*} \and
Shaowei Liu \and
Hanwen Jiang \and \\
Ruihan Yang \and
Yang Fu \and
Xiaolong Wang
}
\authorrunning{Y. Qin et al.}
\institute{University of California San Diego, La Jolla 92093, USA
}
% \end{comment}

%******************
\maketitle
\begin{center}
    \hspace{-2em}
    \centering
       \begin{tabular}{c}
    %\hspace{-0.2in}
    \includegraphics[width=0.8\linewidth]{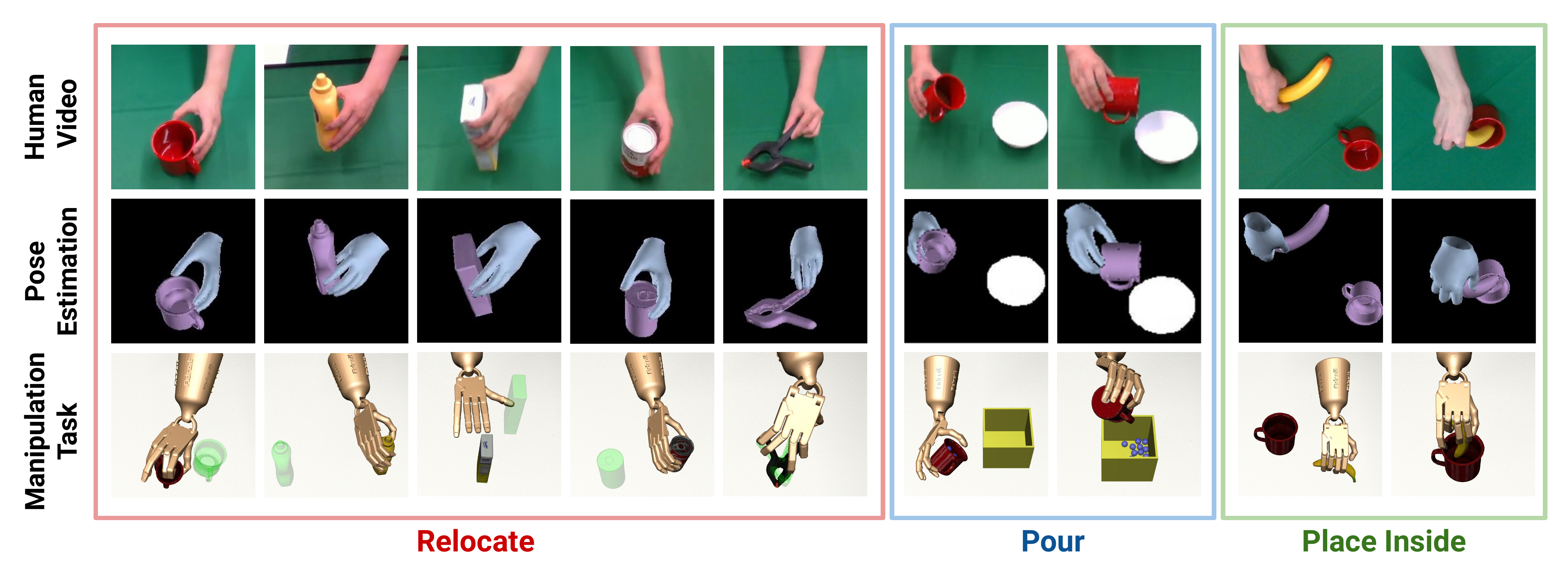}
        \end{tabular}
    \vspace{-0.15in}
    \captionsetup{width=.8\linewidth}
    \captionof{figure}{\small{
    % We propose to perform imitation learning for dexterous manipulation from human demonstration videos. 
    We record human videos on manipulation tasks (1st row) and perform 3D hand-object pose estimations from the videos (2nd row) to construct the demonstrations. We have a paired simulation system providing the same dexterous manipulation tasks for the multi-finger robot (3rd row), including  \emph{relocate}, \emph{pour}, and \emph{place inside}, which we can solve using imitation learning with the inferred demonstrations.}}
    \label{fig:teaser}
    \vspace{-1em}
\end{center}

\begin{abstract}
While significant progress has been made on understanding hand-object interactions in computer vision, it is still very challenging for robots to perform complex dexterous manipulation. In this paper, we propose a new platform and pipeline DexMV (\textbf{Dex}terous \textbf{M}anipulation from \textbf{V}ideos) for imitation learning. We design a platform with: (i) a simulation system for complex dexterous manipulation tasks with a multi-finger robot hand and (ii) a computer vision system to record large-scale demonstrations of a human hand conducting the same tasks. In our novel pipeline, we extract 3D hand and object poses from videos, and propose a novel demonstration translation method to convert human motion to robot demonstrations. We then apply and benchmark multiple imitation learning algorithms with the demonstrations. We show that the demonstrations can indeed improve robot learning by a large margin and solve the complex tasks which reinforcement learning alone cannot solve. More details can be found in the \href{https://yzqin.github.io/dexmv/}{project page}.
\blfootnote{{*} Equal Contribution}
\end{abstract}

\input{sections/introduction}
\input{sections/related_work}
\input{sections/vision}

\input{sections/demonstration}
\input{sections/imitation}
\input{sections/experiment}
\input{sections/conclusion}

% \clearpage
\bibliographystyle{splncs04}
\bibliography{egbib}

\let\cleardoublepage\clearpage
\appendix
\clearpage
% \appendixpage
\input{sections/supp_section}

\end{document}

%% file: sections/introduction.tex
\section{Introduction}
Dexterous manipulation of objects is the primary means for humans to interact with the physical world. Humans perform dexterous manipulation in everyday tasks with diverse objects. To understand these tasks, in computer vision, there is significant progress on 3D hand-object pose estimation~\cite{hasson2019learning,Xiang2018PoseCNNAC} and affordance reasoning~\cite{brahmbhatt2019contactdb,taheri2020grab}. While computer vision techniques have greatly advanced, it is still very challenging to equip robots with human-like dexterity. Recently, there has been a lot of effort on using reinforcement learning (RL) for dexterous manipulation with an anthropomorphic robot hand~\cite{Openai2018}. However, given the high Degree-of-Freedom joints and nonlinear tendon-based actuation of the multi-finger robot hand, it requires a \emph{large amount} of training data with RL. Robot hands trained using only RL will also adopt \emph{unnatural} behavior. Given these challenges, can we leverage humans' experience in the interaction with the physical world to guide robots, with the help of computer vision techniques? 

One promising avenue is imitation learning from human demonstrations~\cite{Rajeswaran2018,schmeckpeper2020reinforcement}. Particularly, for dexterous manipulation, Rajeswaran \etal~\cite{Rajeswaran2018} introduces a simulation environment with four different manipulation tasks and paired sets of human demonstrations collected with a Virtual Reality (VR) headset and a motion capture glove. However, data collection with VR is relatively high-cost and not scalable, and there are only 25 demonstrations collected for each task. It also limits the complexity of the task: It is shown in ~\cite{radosavovic2020state} that RL can achieve similar performance with or without the demonstrations in most tasks proposed in~\cite{Rajeswaran2018}. Instead of focusing on a small scale of data, we look into increasing the difficulty and complexity of the manipulation tasks with diverse daily objects. This requires large-scale human demonstrations which are hard to obtain with VR but are much more available from human videos.

In this paper, we propose \textbf{a new platform and a novel imitation learning pipeline} for benchmarking complex and generalizable dexterous manipulation, namely DexMV  (\textbf{Dex}terous \textbf{M}anipulation from \textbf{V}ideos). We introduce new tasks with the multi-finger robot hand (Adroit Robotic Hand~\cite{kumar2013fast}) on diverse objects in simulation. We collect real human hand videos performing the same tasks as demonstrations. By using human videos instead of VR, it \emph{largely reduces the cost} for data collection and allows humans to perform more \emph{complex and diverse} tasks. While the video demonstrations might not be optimal for perfect imitation (e.g., behavior cloning) to learn successful policies, the diverse dataset is beneficial for augmenting the training data for RL, which can learn from both successful and unsuccessful trials. 

\textbf{Our DexMV platform} contains a paired systems with: (i) A computer vision system which records the videos of human performing manipulation tasks (1st row in Figure~\ref{fig:teaser}); (ii) A physical simulation system which provides the interactive environments for dexterous manipulation with a multi-finger robot (3rd row in Figure~\ref{fig:teaser}). The two systems are aligned with the same tasks. With this platform, our goal is to bridge 3D vision and robotic dexterous manipulation via a novel imitation learning pipeline.

\textbf{Our DexMV pipeline} contains three stages. First, we extract the 3D hand-object poses from the recorded videos (2nd row in Figure~\ref{fig:teaser}). Unlike previous imitation learning studies with 2-DoF grippers~\cite{young2020visual,song2020grasping}, we need the human video to guide the 30-DoF robot hand to move each finger in 3D space. Parsing the 3D structure provides critical and necessary information. Second, a \textbf{key contribution} in our pipeline is a novel demonstration translation method that connects the computer vision system and the simulation system. We propose an optimization-based approach to convert 3D human hand trajectories to robot hand demonstrations. Specifically, the innovations lie in 2 steps: (i) Hand motion retargeting approach to obtain robot hand states; and (ii) Robot action estimation to obtain the actions for learning. Third, given the robot demonstrations, we perform imitation learning in the simulation tasks. We investigate and benchmark algorithms which augment RL objectives with state-only~\cite{radosavovic2020state} and state-action~\cite{Ho2016,Rajeswaran2018} demonstrations. 

We experiment with three types of challenging tasks with the YCB objects~\cite{calli2015benchmarking}. The first task is to \emph{relocate}  an object to a goal position. Instead of relocating a single ball as ~\cite{Rajeswaran2018}, we increase the task difficulty by using diverse objects (first 5 columns in Figure~\ref{fig:teaser}). The second task is \emph{pour}, which requires the robot to pour the particles from a mug into a container (Figure~\ref{fig:teaser} from column 6). The third task is \emph{place inside}, where the robot hand needs to place an object into a container (last 2 columns in Figure~\ref{fig:teaser}). 
In our experiments, we benchmark different imitation learning algorithms and show human demonstrations improve task performance by a large margin. More interestingly, we find the learned policy with our demonstrations can even generalize to unseen instances within the same category or outside the category.  We highlight our contributions as follows:
\begin{itemize}
\vspace{-0.07in}
    \item DexMV platform for learning dexterous manipulation using human videos. It contains paired computer vision and simulation systems with multiple dexterous complex manipulation tasks. 
    \item DexMV pipeline to perform imitation learning, with a key innovation on demonstration translation to convert human videos to robot demonstrations. 
    \item With DexMV platform and pipeline, we largely improve the dexterous manipulation performance over multiple complex tasks, and its generalization ability to unseen object instances.
\end{itemize}

%% file: sections/related_work.tex
\section{Related Work}

\textbf{Dexterous Manipulation.} Manipulation with multi-finger hands is one of the most challenging robotics tasks, which has been actively studied with optimization and planning~\cite{rus1999hand,bicchi2000hands,okamura2000overview,Dogar2010,Andrews2013,Bai2014}. Recently, researchers have started exploring reinforcement learning for dexterous manipulation~\cite{Openai2018,Openai2019}. However, training with only RL requires huge data samples. Different from a regular 2-DoF robot gripper, the Adroit Robotic Hand used in our experiment has 30 DoFs, which greatly increases the optimization space. 

\textbf{Imitation Learning from Human Demonstrations.} Imitation learning is a promising paradigm for robot learning. It is not limited to behavior cloning~\cite{Pomerleau1989,Bain1995,ross2010efficient,Bojarski2016,Torabi2018b} and inverse reinforcement learning ~\cite{Russell1998,Ng2000,Abbeel2004,Ho2016,Fu2017,Aytar2018,Torabi2018g,Liu2020}, but also augmenting the RL training with demonstrations~\cite{Peters2008,Duan2016,Vevcerik2017,Rajeswaran2018,radosavovic2020state}. For example, Rajeswaran \etal~\cite{Rajeswaran2018} propose to incorporate demonstrations with on-policy RL. However, all these approaches rely on expert demonstrations collected using expert policies or VR, which is not scalable and generalizable to complex tasks. In fact, it is shown in~\cite{radosavovic2020state} that RL can achieve similar performance and sample efficiency with or without the demonstrations in most dexterous manipulation tasks in~\cite{Rajeswaran2018} besides \emph{relocate}. Going beyond a VR setup, researchers have recently explored imitation learning and RL with videos~\cite{schmeckpeper2019learning,chang2020semantic,schmeckpeper2020reinforcement,shao2020concept,song2020grasping,young2020visual}. However, all tasks are relatively simple (e.g., pushing a block) with a parallel gripper. In this paper, we propose new challenging dexterous manipulation tasks. We leverage pose estimation for providing demonstrations that largely improve imitation learning performance, while RL alone fails to solve these tasks. 

\textbf{Following Human Demonstrations.} Another line of work is to train policies to follow the expert demonstrations \cite{liu2018imitation,Peng2018s,Pathak2018,Sharma2018,sieb2020graph,garcia2020physics,xiong2021learning,Sermanet2018,sharma19thirdperson,smith2019avid,peng2020learning}. For example, Garcia-Hernando \etal~\cite{garcia2020physics} propose to use RL to followed estimated human hand poses with a virtual robot hand. The robot can execute the same trajectory by following human videos. Instead of repeating one expert trajectory, we emphasize that DexMV is about learning a policy that generalized to different goals and object configurations. 

\textbf{Hand-Object Interaction.} Hand-object interaction is a widely studied topic. One line of work focus on object pose estimation~\cite{Kehl2017SSD6DMR,Rad2017BB8AS,Xiang2018PoseCNNAC,Tekin2018RealTimeSS,Peng2019PVNetPV,Hu2019SegmentationDriven6O,He2020PVN3DAD} and hand pose estimation~\cite{zimmermann2017learning,iqbal2018hand,spurr2018cvpr,ge20193d,baek2019pushing,boukhayma20193d,hasson2019learning,dkulon2020cvpr,liu2021semi}. For example, Berk \etal~\cite{calli2015benchmarking} propose YCB dataset with real objects and their 3D scans. Beyond object poses, datasets and methods for joint estimation of the hand-object poses are also proposed~\cite{GarciaHernando2018FirstPersonHA,hampali2020honnotate,liu2021semi,chao2021dexycb}. Another line of work focus on hand-object contact reasoning~\cite{brahmbhatt2019contactdb,taheri2020grab}, which can be used for functional grasps~\cite{jiang2021graspTTA,brahmbhatt2019contactgrasp,mandikal2020dexterous}. Recently, researchers explored to use hand pose estimation to control a robot hand~\cite{handa2020dexpilot,antotsiou2018task,li2019vision}. They require motion retargeting to map the hand pose into robot motion. These work focus on teleoperation where no objects are involved. In DexMV, we consider the demonstration translation setting, which converts a sequence of hand-object poses into robot demonstrations. We show that the translated demonstration is beneficial for imitation learning for dexterous manipulation.

%% file: sections/vision.tex
\section{Overview}
\label{sec:task}
\vspace{-0.1in}

\begin{figure*}[t]
    \centering
    \includegraphics[width=\linewidth]{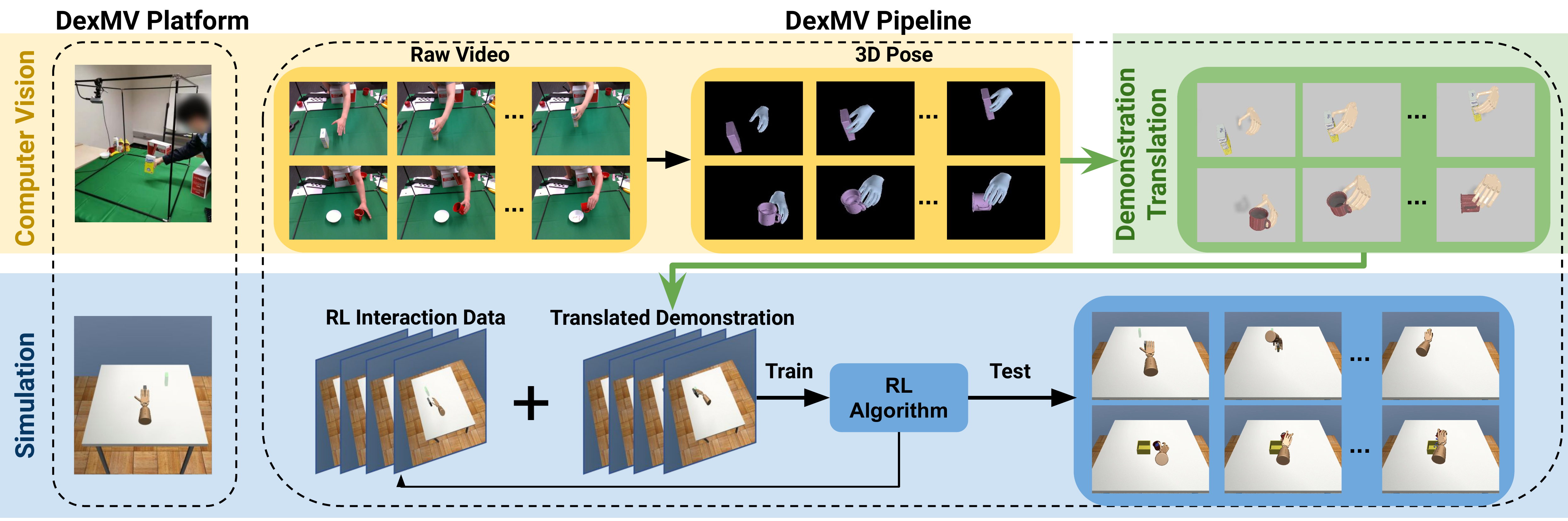}
    \vspace{-0.2in}
    \caption{{\small{\textbf{DexMV platform and pipeline overview.} Our platform is composed of a computer vision system (yellow), a simulation system (blue), and a demonstration translation module (colored with green). In computer vision system, we collect human manipulation videos. In simulation, we design the same tasks for the robot hand. We apply 3D hand-object pose estimation from videos, followed by demonstration translation to generate robot demonstrations, which are then used for imitation learning.}}}
    \vspace{-0.2in}
    \label{fig:main-fig}
\end{figure*}

We propose DexMV for \textbf{Dex}terous \textbf{M}anipulation with imitation learning from human \textbf{V}ideos including: 

\textbf{(i) DexMV platform}, which offers a paired computer vision system recording human manipulation videos and physical simulators conducting the same tasks. 
We design three challenging tasks in the simulator and use computer vision system to collect human demonstrations for these tasks in real world. 
The videos can be efficiently collected with around \emph{100 demonstrations per hour}. 

\textbf{(ii) DexMV pipeline}, a new pipeline for imitation learning from human videos. Given the recorded videos from the computer vision system, we perform pose estimation and motion retargeting to translate human video to robot hand demonstrations. These demonstrations are then used for policy learning where multiple imitation learning and RL algorithms are investigated. By learning from demonstrations, our policy can be deployed in environments with \emph{different goals and object configurations}, instead of just following one trajectory. We show that given the complex manipulation tasks, 3D pose estimation provides \emph{necessary signals} for imitation learning to learn natural policies.

\section{DexMV Platform}
\vspace{-0.1in}
\label{sec:platform}
Our DexMV platform is shown in Figure~\ref{fig:main-fig}. It is composed of a computer vision system and a simulation system. 

\textbf{Computer Vision System.}  The computer vision system is used to collect human demonstration videos on manipulating diverse real objects. In this system, we build a cubic frame (35 inch$^3$) and attach two RealSense D435 cameras (RGBD cameras) on the top front and top left, as shown on the top row of Figure~\ref{fig:main-fig}. During data collection, a human will perform manipulation tasks inside the frame, e.g., relocate sugar box. The manipulation videos will be recorded using the two cameras (from a front view and a side view). 

\textbf{Simulation System.} Our simulation system is built on MuJoCo~\cite{todorov2012mujoco} with the Adroit Hand~\cite{kumar2013fast}. We design multiple dexterous manipulation tasks aligned with human demonstrations. As shown in the bottom row of Figure~\ref{fig:main-fig}, we perform imitation learning by augmenting the Reinforcement Learning with the collected demonstrations. Once the policy is trained, it can be tested on the same tasks with different goals and object configurations. 

\textbf{Task Description.} We propose 3 types of manipulation tasks with different objects. We select the YCB objects~\cite{calli2015benchmarking} with a reasonable size for human to manipulate. We collect 100 demonstrations per object for \emph{relocate} and 100 demonstrations for \emph{pour} and \emph{place inside}. For all tasks, the state is composed of robot joint readings, object pose. For \emph{relocate}, we also include target position. The action is 30-d control command of the position actuator for each joint. 

\emph{Relocate.} It requires the robot hand to pick up an object on the table to a target location. It is inspired by the hardest task in~\cite{Rajeswaran2018} where the robot hand needs to grasp a sphere and move it to the target position. We further increase the task difficulty by using 5 complex objects as shown in the table above and visualized in the first 5 columns of Figure~\ref{fig:teaser}. The transparent green shape represents the goal, which can change during training and testing (goal-conditioned). The task is successfully solved if the object reaches the goal, without the need for a specific orientation. We train one policy for relocating each object.

\emph{Pour.} It requires the robot hand to reach the mug and pour the particles inside into a container (Figure~\ref{fig:teaser} column 6). The robot needs to grasp the mug and then manipulate it precisely. The evaluation criteria is based on the percentage of particles poured inside the container.

\emph{Place Inside.} It requires the robot to pick up an object (e.g., a banana) and place it into a container. The robot needs to rotate the object to a suitable orientation and  approach the container carefully to avoid collisions. 

The evaluation is based on how much percentage of the object mesh volume is inside the container.

Besides testing on the trained objects, we also evaluate the generalization ability of the trained policies on unseen object instances, within the training categories like can, bottle, mug, and also outside the training categories such as camera from ShapeNet dataset~\cite{chang2015shapenet}.

\vspace{-0.05in}
\section{Pose Estimation}
\label{sec:pose_estimation}

\vspace{-0.05in}
\subsection{Object Pose Estimation}
\label{sec:object_pose}
\vspace{-0.05in}
We use the 6-DoF pose to represent the location and orientation of objects, which contains translation $T \in \mathbf{R}^3$ and rotation $R \in \mathbf{SO(3)}$. For each frame $t$ in the video, we use the PVN3D~\cite{He2020PVN3DAD} model trained on the YCB dataset~\cite{calli2015benchmarking} to detect objects and estimate 6-DoF poses. By taking both the RGB image and the point clouds as inputs, the model first estimates the instance segmentation mask. With dense voting on the segmented point clouds, the model then predicts the 3D location of object key points. The 6-DoF object pose is optimized by minimizing the PnP matching error. 

\subsection{Hand Pose Estimation}
\label{sec:hand_pose}
\vspace{-0.05in}
We utilize the MANO model~\cite{romero2017embodied} to represent the hand that consists of hand pose parameter $\theta_t$ for 3D rotations of 15 joints and root global pose $r_t$, and shape parameters $\beta_t$ for each frame $t$. The 3D hand joints can be computed using hand kinematics function $j_t^{3d} = \mathbf{J}(\theta_t, \beta_t, r_t)$. 

Given a video, we use the off-the-shelf skin segmentation~\cite{khaled2009combinatorial} and hand detection~\cite{Shan20} models to obtain a hand mask $M_t$. We use the trained hand pose estimation models in~\cite{liu2021semi} to predict the 2D hand joints $j_t^{2d}$ and the MANO parameters $\theta_t$ and $\beta_t$ for every frame $t$ using RGB image. We estimate the root joint $r_t$ using the center of the depth image masked by $M_t$. Given the initial estimation $\theta_t, \beta_t, r_t$ of each frame $t$ and the camera pose $\Pi$, we formulate the 3D hand joint estimation as an optimization problem,

\begin{equation}
\label{hand-opt}
\begin{split}
\theta_t^{*}, \beta_t^{*}, r_t^{*} = \argmin_{\theta_t, \beta_t, r_t} {\vert\vert \Pi J(\theta_t, \beta_t, r_t) - j_t^{2d} \vert\vert^2 +\lambda \vert\vert M_t \cdot (\mathbf{R}(\theta_t, \beta_t, r_t) - D_t)\vert\vert^2} \\
\end{split}
\end{equation}
where $\lambda=0.001$ and $\mathbf{R}$ is a depth rendering function~\cite{kato2018neural} and $D_t$ is the corresponding depth map in frame $t$. We minimize the re-projection error in 2D and optimize the hand 3D locations from the depth map. Equation~\ref{hand-opt} can be further extended to a multi-camera setting by minimizing the objective from different cameras with calibrated extrinsics. We use two cameras by default to handle occlusion. Additionally, we deploy a post-processing procedure with minimizing the difference of $j_{t}^{3d}$ and $\beta_t$ between frames.

%% file: sections/demonstration.tex
\vspace{-0.05in}
\section{Demonstration Translation}
\label{sec:demo_translation}
\vspace{-0.05in}
Common imitation learning algorithm consumes state-action pairs from expert demonstrations. It requires the state of the robot and the action of the motor as training data but not directly using the human hand pose. As shown in the kinematics chains in the blue box of Figure~\ref{fig:retargeting}, although both robot and human hands share a similar five-finger morphology, their kinematics chains are different. The human hand MANO model~\cite{romero2017embodied} is parameterized by $15$ rotation vector, leading to $15$ ball joint with $15 *3 = 45$ DoF. The Adroit Robotic Hand~\cite{kumar2013fast} used in simulation has $24$ revolute joint and $1$ free joint, which leads to $24 + 1*6=30$ DoF. The finger length of each knuckle is also different.

In this paper, we propose a novel method to translate demonstrations from \emph{human-centric pose estimation} result into \emph{robot-centric imitation data}. Specifically, there are two steps in demonstration translation (as shown in the red box of Figure~\ref{fig:retargeting}): (i) \textbf{Hand motion retargeting} to align the human hand motion to robot hand motion, which are with different DoF and geometry; (ii) \textbf{Predicting robot action}, i.e. torque of robot motor: Without any wired sensors, we need to recover the action from only pose estimation results. We will introduce our approach for these two challenges as follows. 

\vspace{-0.05in}
\subsection{Hand Motion Retargeting}
\label{sec:retarget}
\vspace{-0.05in}
\begin{figure}[!t]
%\vspace{-0.3in}
\centering
\begin{minipage}{.42\textwidth}
  \centering
  \vspace{-0.1in}
  \includegraphics[width=0.9\linewidth]{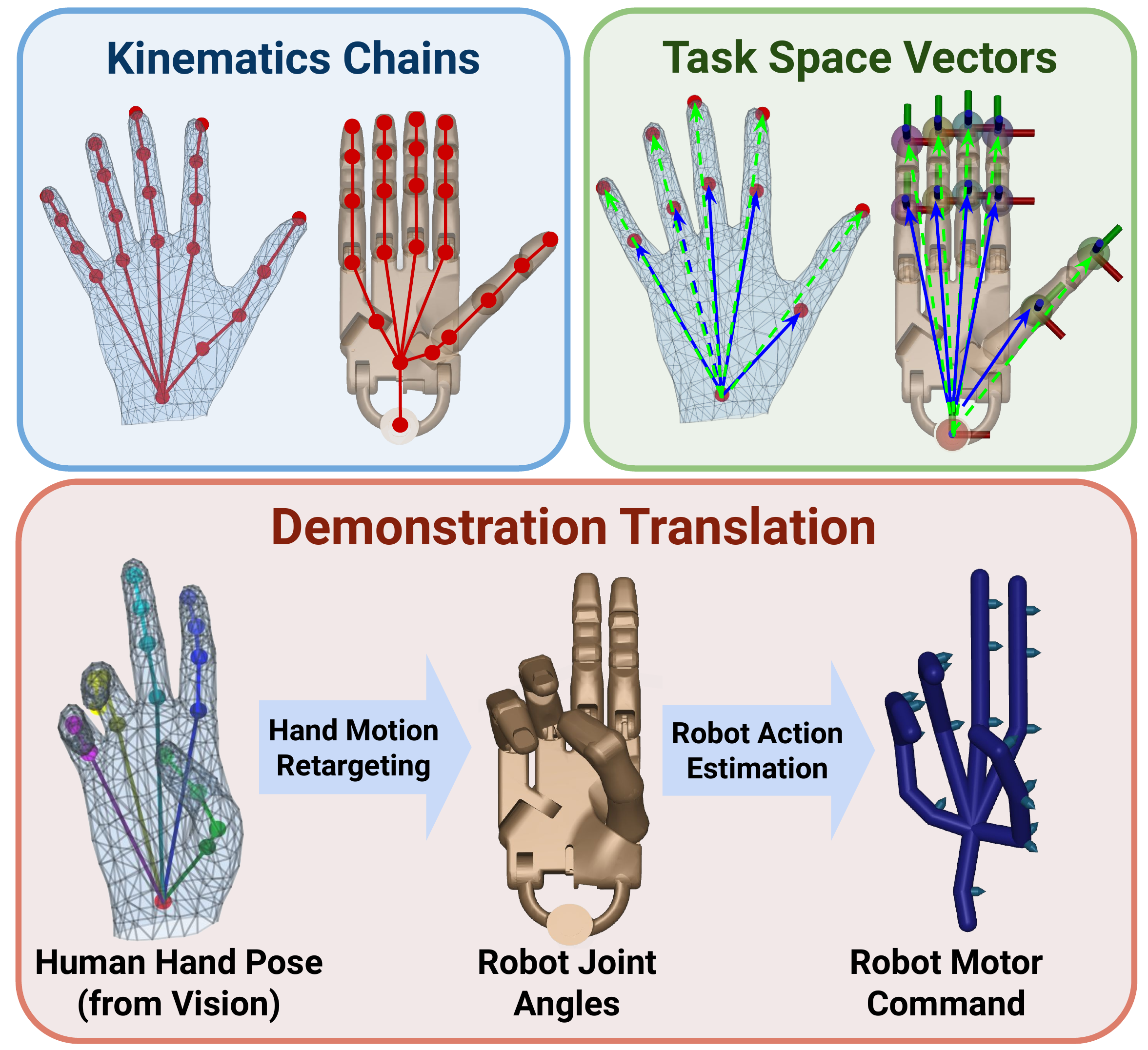}
  \captionsetup{width=\linewidth}
  \caption{\small{\textbf{Top row}: Kinematic Chains and Task Space Vectors (TSV). The TSV (dash arrows) are ten vectors to be matched between both hands. \textbf{Bottom row}: (i) Hand motion retargeting; (ii) Action Estimation.}} 
  \label{fig:retargeting}
\end{minipage}\hfill
\begin{minipage}{.54\textwidth}
  \centering
  \vspace{-0.01in}
  \includegraphics[width=0.9\linewidth]{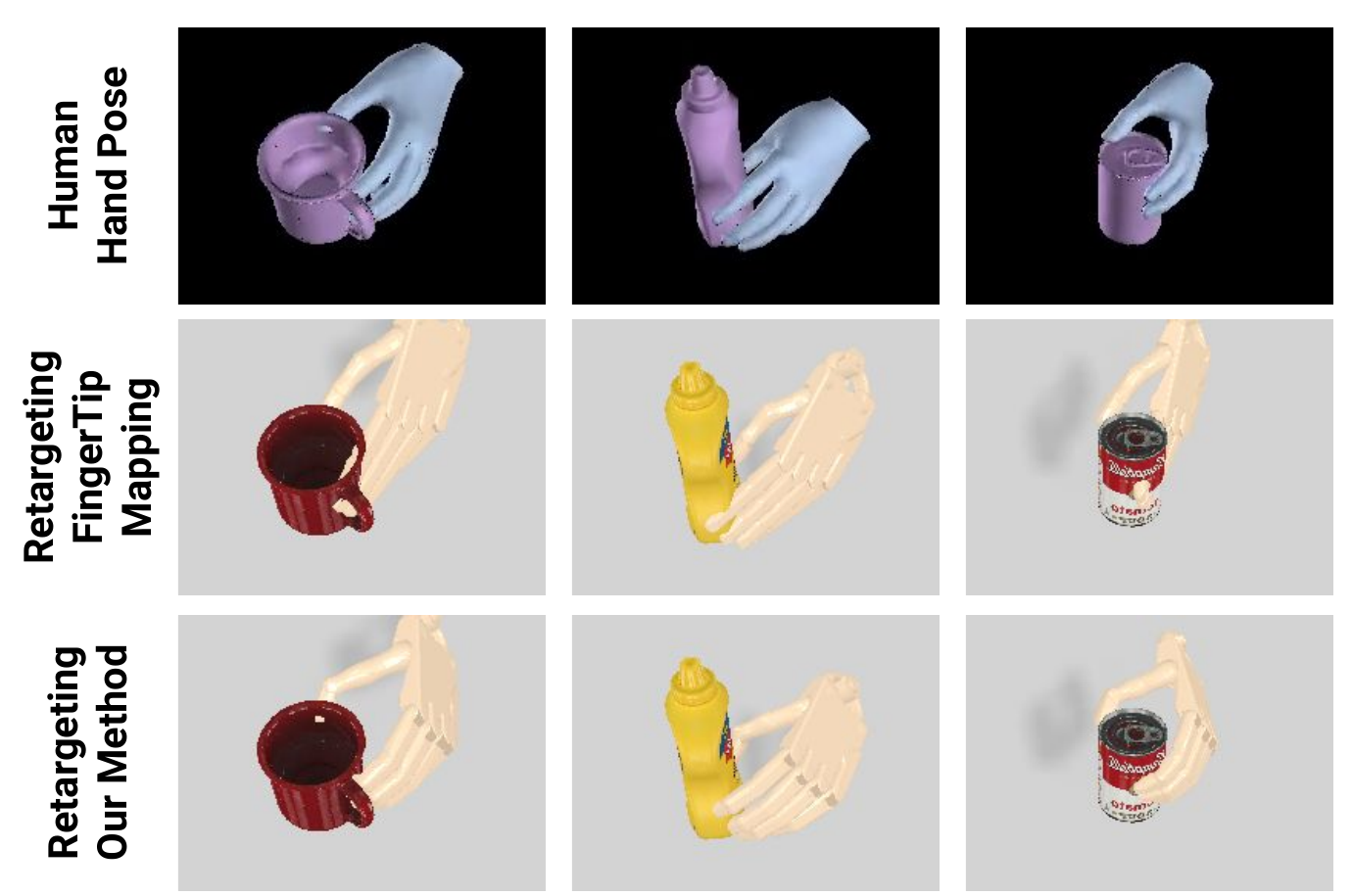}
  \captionsetup{width=\linewidth}
  \caption{\small{Visualization of hand motion retargeting results. \textbf{Top row:} 3D hand-object poses. \textbf{Mid row:} Retargeting results using Finger Tip Mapping as optimization objective. \textbf{Bottom row:} Retargeting results using the proposed optimization objective.}}
  \label{fig:retargeting_result}
\end{minipage}
\vspace{-2em}
\end{figure}

In computer graphics and animation, motion retargeting is used to retarget a performer's motion to a virtual character for applications such as movies and cartons~\cite{aberman2019learning,hecker2008real}. While the animation community emphasizes realistic visual appearance, we consider more on the physical effect of the retargeted motion in the context of robotics manipulation. That is, the robot motion should be executable, respecting the physical limit of motors, e.g. acceleration and torque.

Formally, given a sequence of human hand pose estimated $\{(\theta_t, \beta_t, r_t)\}_{t=0}^{T}$ from a video, the hand pose retargeting can be defined as computing the sequence of robot joint angles $\{q_t\}_{t=0}^{T}$, where $t$ is the step number in a sequence of length $T+1$. We formulate the hand pose retargeting as an optimization problem.

\textbf{Optimization with Task Space Vectors.} 
In most manipulation tasks, human and multi-finger robot hand contacts the object using finger tips. Thus in fingertip-based mapping~\cite{antotsiou2018task,handa2020dexpilot}, retargeting is solved by preserving the Task Space Vectors (TSV) defined from finger tips position to the hand palm  position (green arrow) as shown in the green box of Figure~\ref{fig:retargeting}, so that the human and robot hands will have same finger tip position relative to the palm. However, only considering finger tip may results in unexpected optimization result as shown in Figure~\ref{fig:retargeting_result}. Although the tip position is preserved, the bending information of hand fingers are lost, leading to penetrating the object after retargeting. It becomes more severe when the joint angles are closed to the robot singularity~\cite{nakamura1986inverse}. To solve this issue, we include the TSV from palm to middle phalanx (blue arrow) as shown in the green box of Figure~\ref{fig:retargeting}. The optimization objective is, 

\vspace{-0.7em}
\begin{equation}
\vspace{-0.5em}
\label{eq:retargeting_final}
\min_{q_t} \sum_{i=0}^{N} || \mathbf{v_i^H}(\theta_t, \beta_t, r_t) - \mathbf{v_i^R}(q_t) ||^2 + \alpha||q_t - q_{t-1}||^2,
\end{equation}

where the forward kinematics function for human $\mathbf{v_i^H}(\theta, \beta, r)$ computes the $i$-th TSV for human hand and the robot forward kinematics function $\mathbf{v_i^R}(q)$ computes $i$-th TSV for the robot. The first term in Equation~\ref{eq:retargeting_final} is to find the best $\{q_t\}_{t=0}^{T}$ that matches the finger tip-palm TSV and finger tip-phalanx TSV from both hands.  Note the human hand TSV $\{v_i^H(\theta_0, \beta_0, r_0)\}_{t=0}^{T}$ is first processed by a low pass filter before optimization for better smoothness. We also add an L2 normalization (second term in Equation~\ref{eq:retargeting_final}) to improve the temporal consistency. For optimization, we use Sequential Least-Squares Quadratic Programming in NLopt~\cite{johnson2014nlopt}, where $\alpha=8e-3$ in  implementation.

When retargeting hand pose, the objective in Equation~\ref{eq:retargeting_final} is optimized for each $t$ from $t=0$ to $t=T$. For $t \ge 1$, we initialize $q_t$ using the optimization results $q_{t-1}$ from last step to further improve the temporal smoothness.

\vspace{-0.05in}
\subsection{Robot Action Estimation}
\label{sec:time_param}
\vspace{-0.05in}
Hand motion retargeting provides temporal-consistent translation from human hand poses to robot joint angles $\{q_t\}_{t=0}^{T}$ in different time steps. But the action, i.e. joint torque or control command, is unknown. In robotics, the joint torque $\tau$ can be computed via robot inverse dynamics function $\tau=f_{inv}(q, q', q'')$. To use this function, we first fit the sequence of joint angles into a continuous joint trajectory function $\mathbf{q}(t)$. Then, the first and second order derivative $\mathbf{q}'(t)$ and $\mathbf{q}''(t)$ can be used to compute the torque $\tau(t) = f_{inv}(q(t), q'(t), q''(t))$.

\textbf{One important property of $\mathbf{q}(t)$ is:}  The absolute value of third-order derivative $\mathbf{q}'''(t)$, which also refers as jerk, should be as small as possible. We call it minimum-jerk requirement. The reason is two-fold: (i) First, physiological study~\cite{flash1985coordination} shows that the motion of human hand is a minimum jerk trajectory, where the lowest effort is required during the movement. This requirement will resemble the behavior of human, which in turn leads to more natural robot motion. (ii) Second, minimizing jerk can ensure low joint position errors for motors~\cite{kyriakopoulos1988minimum} and limit excessive wear on the physical robot~\cite{craig2009introduction}. Thus we also expect the fitted trajectory function $\{q_t\}_{t=0}^{T}$ to be minimum-jerk. In implementation, we use the model proposed in \cite{todorov1998smoothness} to achieve this.

\textbf{Time Alignment.} The recorded video is with around $30Hz$ frequency and the simulation runs at $120 Hz$. We need to align the demonstrations with the simulated environment before we apply for training. We sample the continuous function $\mathbf{q}(t)$ with simulation frequency and compute the corresponding action. 

%% file: sections/imitation.tex
\vspace{-0.05in}
\section{Imitation Learning}
\label{sec:imlearn}
\vspace{-0.05in}
We perform imitation learning using the translated demonstrations. Instead of using behavior cloning, we adopt imitation learning algorithms that incorporate the demonstrations into RL. 

\textbf{Background.}  
We consider the Markov Decision Process (MDP) represented by $\langle S, A, P, R,\gamma\rangle$, where $S$ and $A$ are state and action space, $P(s_{t+1}\vert s_t,a_t)$ is the transition density of state $s_{t+1}$ at step $t+1$ given action $a_t$. $R(s,a)$ is the reward function, and $\gamma$ is the discount factor. The goal of RL is to maximize the expected reward under policy $\pi(a\vert s)$.  We incorporate RL with imitation learning. Given trajectories $\{(s_i,a_i)\}_{i=1}^n$ from demonstrations $\pi_\mathrm{D}$, we optimize the agent policy $\pi_\theta$ with $\{(s_i,a_i)\}_{i=1}^n$ and reward $R$. We evaluate imitation learning under two settings:  state-action imitation and state-only imitation. All methods utilize both demonstrations and reward to learn the dexterous manipulation tasks. 

\vspace{-0.05in}
\subsection{State-action imitation learning} 
\vspace{-0.05in}
We will introduce two algorithms. The first one is the  Generative Adversarial Imitation Learning (GAIL)~\cite{ho2016generative}, which is the SOTA IL method that performs occupancy measure matching to learn policy. Occupancy measure $\rho_\pi$~\cite{Puterman:1994:MDP:528623} is a density measure of the state-action tuple on policy $\pi$. GAIL aims to minimize the objective $d(\rho_{\pi_\mathrm{E}},\rho_{\pi_\theta})$, where $d$ is a distance function, $\pi_\theta$ is the policy parameterized by $\theta$. 
GAIL use generative adversarial training to estimate the distance and minimize it with the objective as,

\vspace{-1em}
\begin{align}
    \scalebox{.8}{$\underset{\theta}{\min}\underset{w}{\max}$}
    \underset{{\scalebox{.6}{$s,a\sim \rho_{\pi_\theta}$}}}{\mathbb{D}}[\log D_w(s,a)]+\underset{{\scalebox{.6}{$s,a\sim \rho_{\pi_\mathrm{D}}$}}}{\mathbb{E}}[\log (1-D_w(s,a))],
    \vspace{-8mm}
\end{align}
where $D_w$ is a discriminator.
To equip GAIL with reward function, we adopt the approach proposed by~\cite{kang2018policy}, denoted as GAIL+ in this paper.

The second algorithm is Demo Augmented Policy Gradient (DAPG)~\cite{rajeswaran2017learning}. 
The objective function for policy optimization at each iteration $k$ is as follow.
\vspace{-1em}
\begin{align}%\label{eq:dapg}
    g_{aug}=\sum_{(s,a)\in\rho_{\pi_\theta}}\nabla_\theta \ln\pi_\theta(a\vert s) A^{\pi_\theta} (s,a)+\sum_{(s,a)\in\rho_{\pi_\mathrm{D}}}\nabla_\theta \ln \pi_\theta (a\vert s)\lambda_0\lambda_1^k, 
\end{align}
where $A^{\pi_\theta}$ is the advantage function~\cite{baird1993advantage} of policy $\pi_\theta$, and $\lambda_0$ and $\lambda_1$ are hyper-parameters to determine the advantage of state-action pair in demonstrations.

\vspace{-0.05in}
\subsection{State-only imitation learning} 
\vspace{-0.05in}
Besides state-action imitation, we also evaluate the State-Only Imitation Learning (SOIL)~\cite{radosavovic2020state} algorithm which does using action information from demonstrations. SOIL extends DAPG to the state-only imitation setting by learning an inverse model $h_\phi$ with the collected trajectories when running the policy.
The inverse model predicts the missing actions in demonstrations and the policy is trained as DAPG. The policy and inverse model are optimized simultaneously.

%% file: sections/experiment.tex
\vspace{-0.2in}
\section{Experiment}
\vspace{-0.05in}
\label{sec:exp}

We conduct experiments on the proposed tasks including \emph{Relocate}, \emph{Pour} and \emph{Place Inside} defined in Section~\ref{sec:task}. We benchmark different imitation learning algorithms on two aspects: (i) First, we benchmark the different methods by testing on the trained objects but with different configurations. We report the training curves and success rates for all tasks. We also ablate how different ways for hand pose estimation, number of demonstrations, and different environmental parameters can affect imitation learning. (ii) Second, we benchmark how well different imitation learning algorithms can generalize to unseen object instances. We evaluate on both object instances that are taken from the same category as the trained object, and from a different category. 

\textbf{Experiment settings.} We adopt TRPO~\cite{schulman2015trust} as our RL baseline. We benchmark imitation learning algorithms SOIL, GAIL+, and DAPG. All these algorithms incorporate demonstrations with TRPO of same hyper-parameters. Thus all the algorithms are comparable. 
All policies are evaluated with the same three random seeds. By default, the hand pose is estimated using two cameras. 

\vspace{-0.1in}
\subsection{Experiments with Relocate}
\begin{figure*}[!t]
    \centering
    \includegraphics[width=1\linewidth]{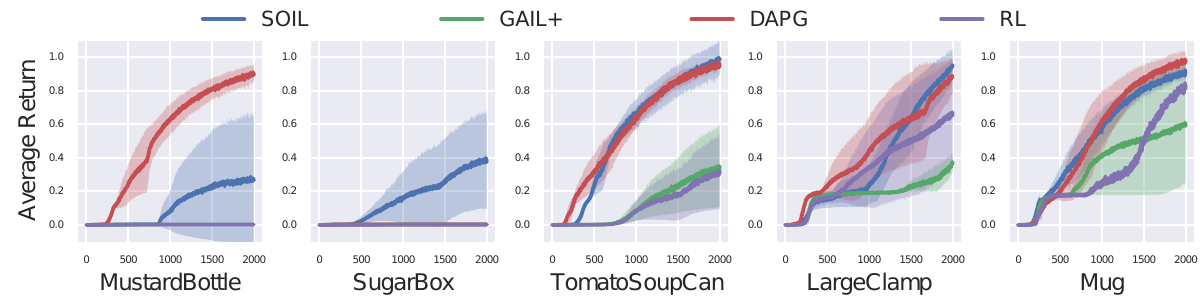}
    \vspace{-0.25in}
    \caption{\small{Learning curves of the four methods on the relocate task with respect to five different objects. The x-axis is training iterations. The shaded area indicates standard error and the performance is evaluated with three individual random seeds. }}
    \label{fig:relocation}
\end{figure*}

\vspace{-0.05in}
\textbf{Main comparisons.} We benchmark four methods: SOIL, GAIL+, DAPG, and RL on the \emph{relocate} tasks. The \emph{success rate} is shown in Table~\ref{tab:success_rate} and training curves are in Figure~\ref{fig:relocation}. A trial is counted as success only when the final position of the object (after 200 steps) is within 0.1 unit length to the target. The initial object position and target position are randomized. In figure~\ref{fig:relocation}, the x-axis is training iterations and the y-axis is normalized average-return over three seeds. 

Both Table~\ref{tab:success_rate} and Figure~\ref{fig:relocation} show that the imitation learning methods outperform RL baseline for all five tasks. For mustard bottle and sugar box, a pure RL agent is not able to learn anything. SOIL performs the best for sugar box while DAPG achieves comparable or better performance on mug, mustard bottle, and tomato soup can. \emph{Relocate} with sugar box is most challenging since it is tall and thin with a flat surface. Once it drops on the table, it is hard to grasp again. We conjecture training the inverse dynamics model online in SOIL helps obtain more accurate action for \emph{Relocate}. 

\begin{table*}[t]
    \tiny
    \centering
    \vspace{-0.1in}
    \begin{minipage}[t]{1\linewidth}
    \tablestyle{5pt}{1.05}
    \resizebox{\textwidth}{!}{%
    \begin{tabular}{l|c|c|c|c|c}
         & \multicolumn{5}{c}{Task - Relocate} \\
        Model & Mustard & Sugar Box & Tomato Can & Clamp & Mug \\\shline
        SOIL & $0.33\pm 0.42$ & $\textbf{0.67}\pm \textbf{0.47}$ & $0.98 \pm 0.02$ & $0.89\pm0.15$ & $0.71\pm0.35$\\
        GAIL+ & $0.06\pm0.01$ & $0.00\pm0.00$ & $0.66\pm0.47$ &$0.52\pm0.39$& $0.53\pm0.37$ \\
        DAPG & $\textbf{0.93}\pm\textbf{0.05}$ & $0.00\pm0.00$& $\textbf{1.00}\pm\textbf{0.00}$& $\textbf{1.00}\pm\textbf{0.00}$ & $\textbf{1.00}\pm\textbf{0.00}$ \\
        RL & $0.06\pm0.01$ & $0.00\pm0.00$ & $0.67\pm0.47$ &$0.51\pm0.37$& $0.49\pm0.36$ \\
    \end{tabular}}
    \vspace{0.05in}
    \caption{\small{Success rate of the evaluated methods on \emph{Relocate} with five different objects. Success is defined based on the distance between object and target, evaluated via $100$ trials for three seeds. }}
    \label{tab:success_rate}
    \end{minipage}
\vspace{-3em}
\end{table*}

\begin{figure*}[t]
% Fig 1
    \begin{minipage}{0.24\textwidth}
    \centering
    \includegraphics[width=\textwidth]{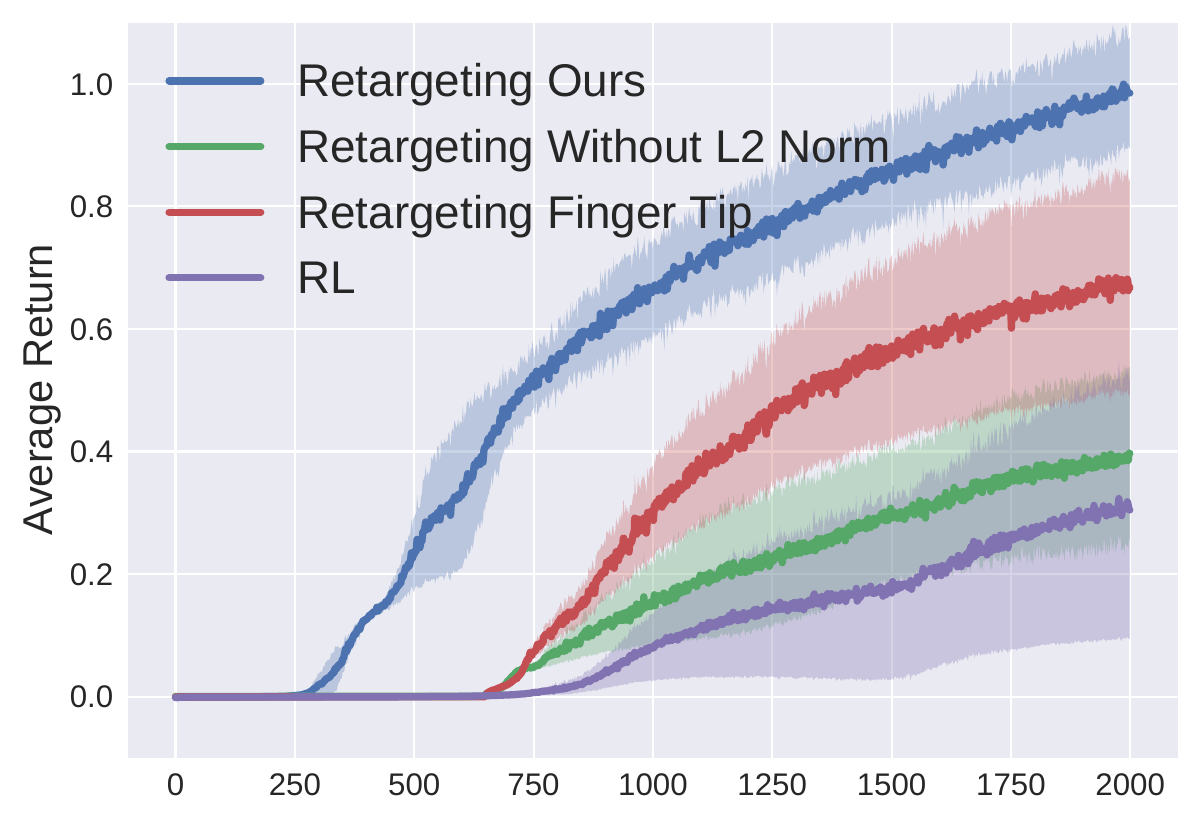} \\
    \quad \small{(a) Retargeting }
	\end{minipage}
% Fig 2
    \begin{minipage}{0.24\textwidth}
    \centering
    \includegraphics[width=\textwidth]{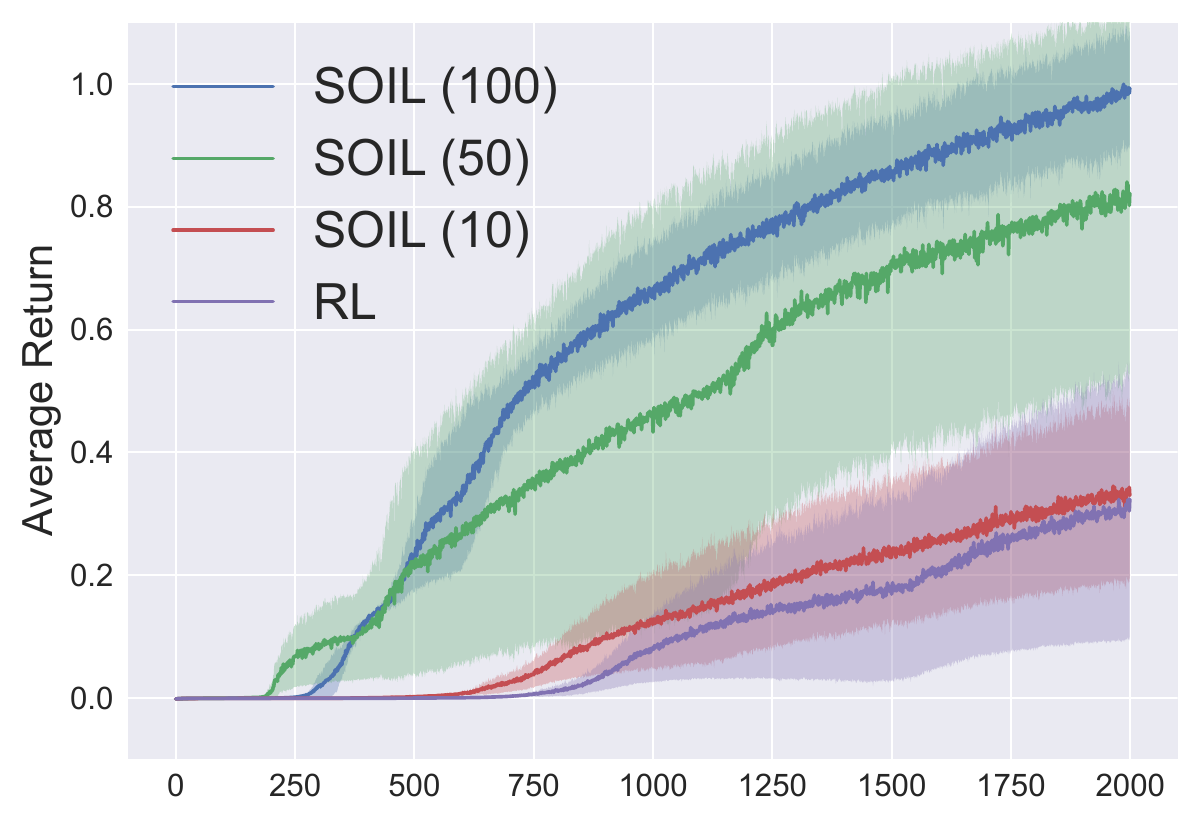} \\
    \quad \small{(b) Num of Demo.}
	\end{minipage}
% Fig 3
	\begin{minipage}[h]{0.24\linewidth}
% 	\vspace{-0.9em}
    \centering
    \includegraphics[width=\textwidth]{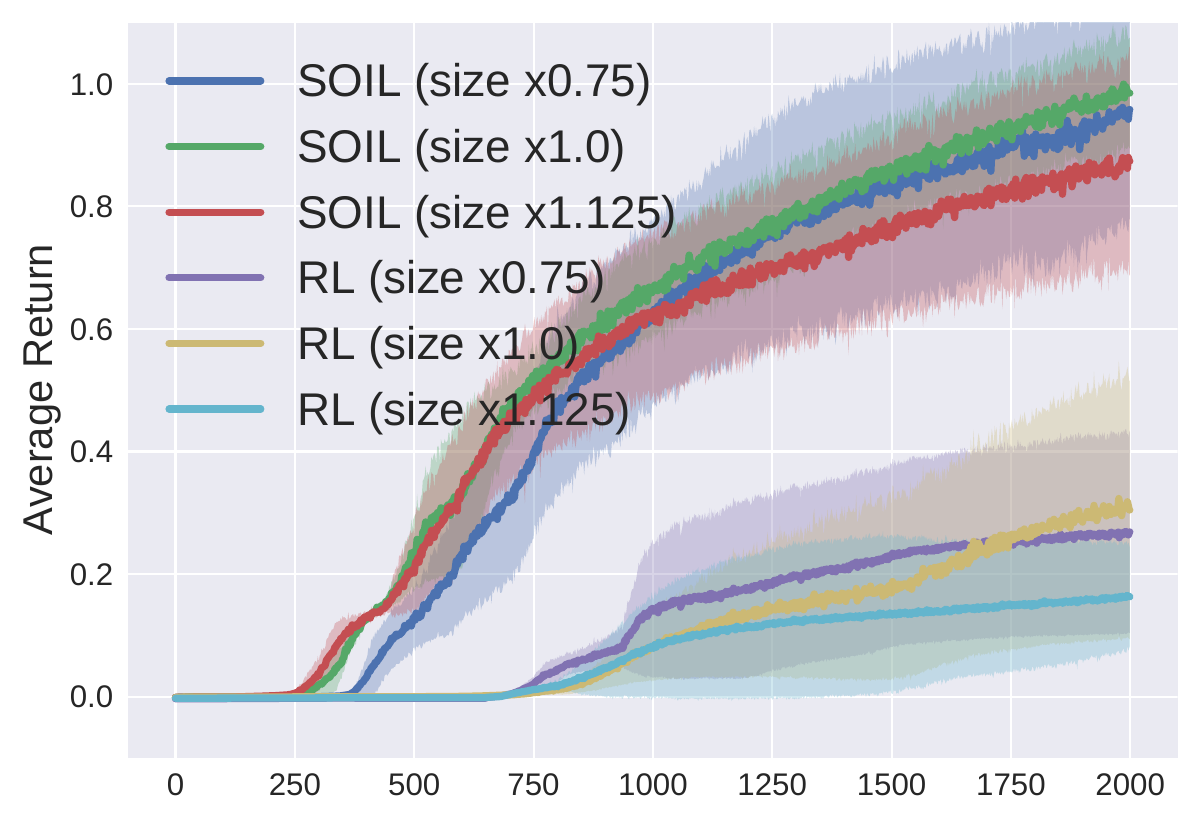} \\
    \quad \small{(c) Object Size}
    \end{minipage}
% Fig 4
	\begin{minipage}[h]{0.24\linewidth}
% 	\vspace{-0.9em}
    \centering
    \includegraphics[width=\textwidth]{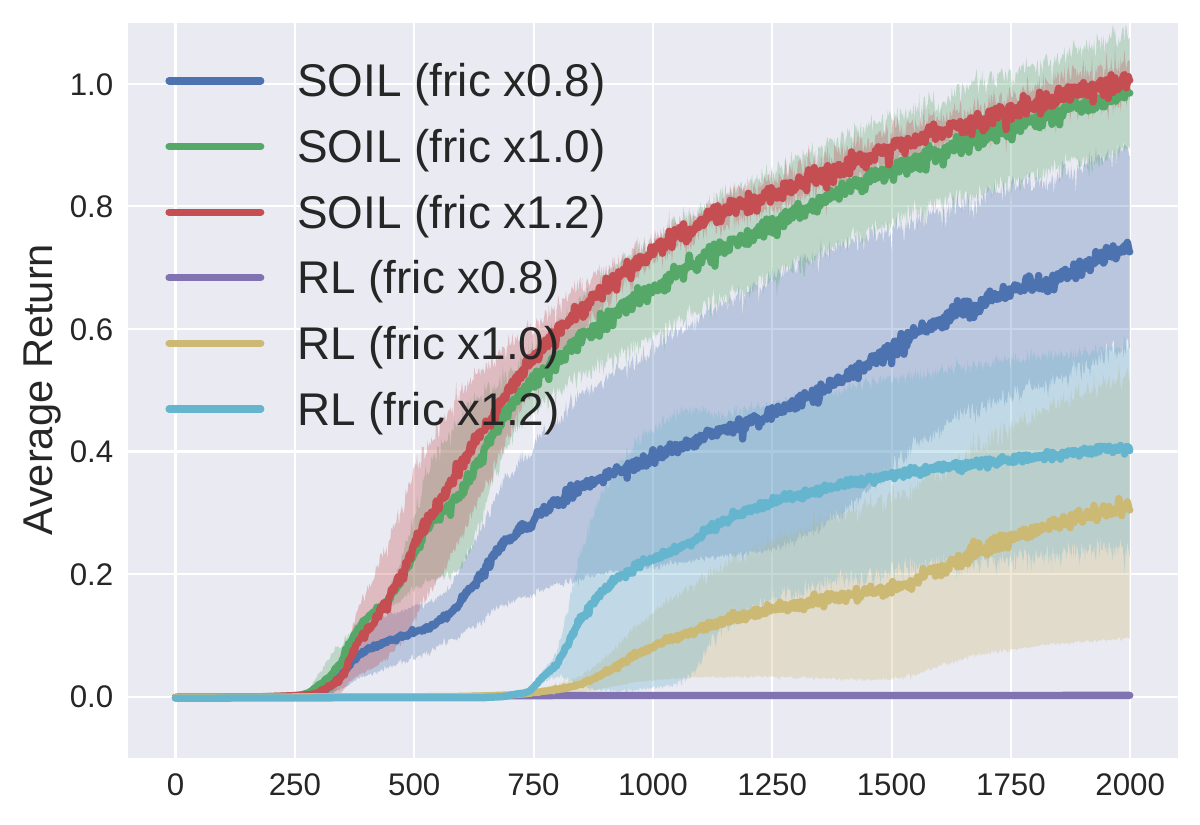}
    \quad  \small{(d) Object Friction}
    \end{minipage}%
    \vspace{-0.5em}
    \caption{\small{\textbf{Ablation Study}: Learning curves of SOIL on \emph{Relocate} with tomato soup can. The x-axis is training iterations. We ablate: (a) hand pose estimation methods; (b) number of demonstrations used to train SOIL; (c) scaling of object size; (d) friction of the relocated object. The demonstrations are kept the same for all conditions.}}
    \label{fig:ablation}
    % \vspace{-2em}
\end{figure*}
\begin{table}[t]
    \begin{minipage}{0.55\linewidth}
    \vspace{-0.2em}
    \resizebox{0.95\columnwidth}{!}{%
    \begin{tabular}{c|c|c|c}
            $\#$Demo & 400 Iter. & 600 Iter. & 800 Iter. \\\shline
            SOIL (100) & $\textbf{0.36}\pm\textbf{0.13}$ & $\textbf{0.70}\pm\textbf{0.25}$ & $\textbf{1.00}\pm\textbf{0.00}$\\
            SOIL (50) &  $0.19\pm0.23$ & $0.40\pm0.43$ & $0.59\pm0.39$\\
            SOIL (10) & $0.04\pm0.05$ & $0.17\pm0.11$  & $0.36\pm 0.21$\\
            RL (0) &  $0.00\pm0.00$ & $0.00\pm0.00$ & $0.13\pm0.19$\\
    \end{tabular}}
    \vspace{0.5em}
    \captionsetup{width=0.95\linewidth}
    \caption{\small{Success rate with different number of demonstrations on \emph{Relocate} task with a tomato soup can, evaluated via $100$ trials for three random seeds at 400, 600, and 800 training iterations.}}
    \label{tab:num_success_rate}
    \end{minipage} \hfill
    \begin{minipage}{0.46\linewidth}
    \centering
    \vspace{-0.1em}
    \resizebox{0.8\columnwidth}{!}{%
    \begin{tabular}{c|c|c}
    Setting & MPJPE & Success (\%) \\ \shline
    1 Cam & $41.7$ & $66.7 \pm 57.7$ \\
    1 Cam Post & $36.2$ & $69.7 \pm 33.3$ \\
    2 Cam & $36.6$ & $84.7 \pm 25.7$ \\
    2 Cam Post & $\textbf{32.5}$ & $\textbf{93.3} \pm \textbf{11.5}$ \\
    \end{tabular}}
    \vspace{0.3em}
    \captionsetup{width=0.9\linewidth}
    \captionof{table}{\small{Hand pose error and the success rate of learned policy for 4 hand pose estimation settings. 
    Smaller MPJPE means better hand pose estimation performance.
    }}
    \label{tab:hand_pose}
    \end{minipage}
    \vspace{-2em}
\end{table}

\vspace{0.05in}
\textbf{Ablation on motion retargeting method.} 
Figure~\ref{fig:ablation} (a) illustrates the performance of SOIL with respect to demonstrations generated by different retargeting method, on relocating a tomato soup can. It shows that our retargeting design (blue curve) can improve the imitation learning performance over fingertip mapping (red curve) based retargeting baseline. Besides, without L2 norms in Equation~\ref{eq:retargeting_final} (green curve), the performance shows a huge drop. It indicates the importance of smooth robot motion when used as demonstrations.

\vspace{0.05in}
\textbf{Ablation on the number of demonstrations.} Figure~\ref{fig:ablation} (b) shows the performance of SOIL with respect to different number of demonstrations, on relocating a tomato soup can. SOIL can achieve better sample efficiency, performance and the variance is reduced when more demonstrations are given. Table~\ref{tab:num_success_rate} illustrates the success rate of the different policies trained with a different number of demonstrations. \emph{This shows the importance of our efficient platform to scale up the number and diversity of demonstrations.}

\vspace{0.05in}
\textbf{Ablation on environmental conditions.}
Figure~\ref{fig:ablation} (c) and (d) illustrates that our demonstrations can transfer to different physical environments.  Given the same demonstrations on \emph{Relocate} the tomato soup can, we train policies with objects scaling into different size from $\times0.75$ to $\times1.125$ (Figure~\ref{fig:ablation} (c)) and applying different frictions (Figure~\ref{fig:ablation} (d)) from $\times0.8$ to $\times1.2$. With same demonstrations, we can still perform imitation learning in different environments and achieve consistent results. All policies achieve much better performance than pure RL. \emph{This experiment shows the robustness of our pipeline against the gap between simulation (environment) and real (demonstration).}

\vspace{0.05in}
\textbf{Ablation on hand pose estimation.} 
We choose 4 different settings for hand pose estimation from video captured by: (i) single camera; (ii) single camera plus post-processing (iii) dual camera; (iv) dual cameras plus the post-processing (mentioned in Section~\ref{sec:hand_pose})
Since we have no ground-truth pose annotation in our dataset, we evaluate the performance of hand pose estimation approaches on the DexYCB~\cite{chao2021dexycb} dataset, which provides the pose ground-truths. We follow DexYCB to use Mean Per Joint Position Error (MPJPE, smaller the better). We experiment with SOIL on relocating a tomato soup can. The object pose remains the same for all four settings. Table~\ref{tab:hand_pose} shows better pose estimation in general corresponds to better imitation, except the comparison on 1-Cam Post and 2-Cam. It seems 2-Cam are better than 1-Cam in imitation, even pose estimation result is close. We conjecture 2-Cam can provide more smooth trajectories. 
\vspace{-1em}

\begin{table}[!t]
	\begin{minipage}{0.27\linewidth}
		\centering
		\includegraphics[width=0.95\linewidth]{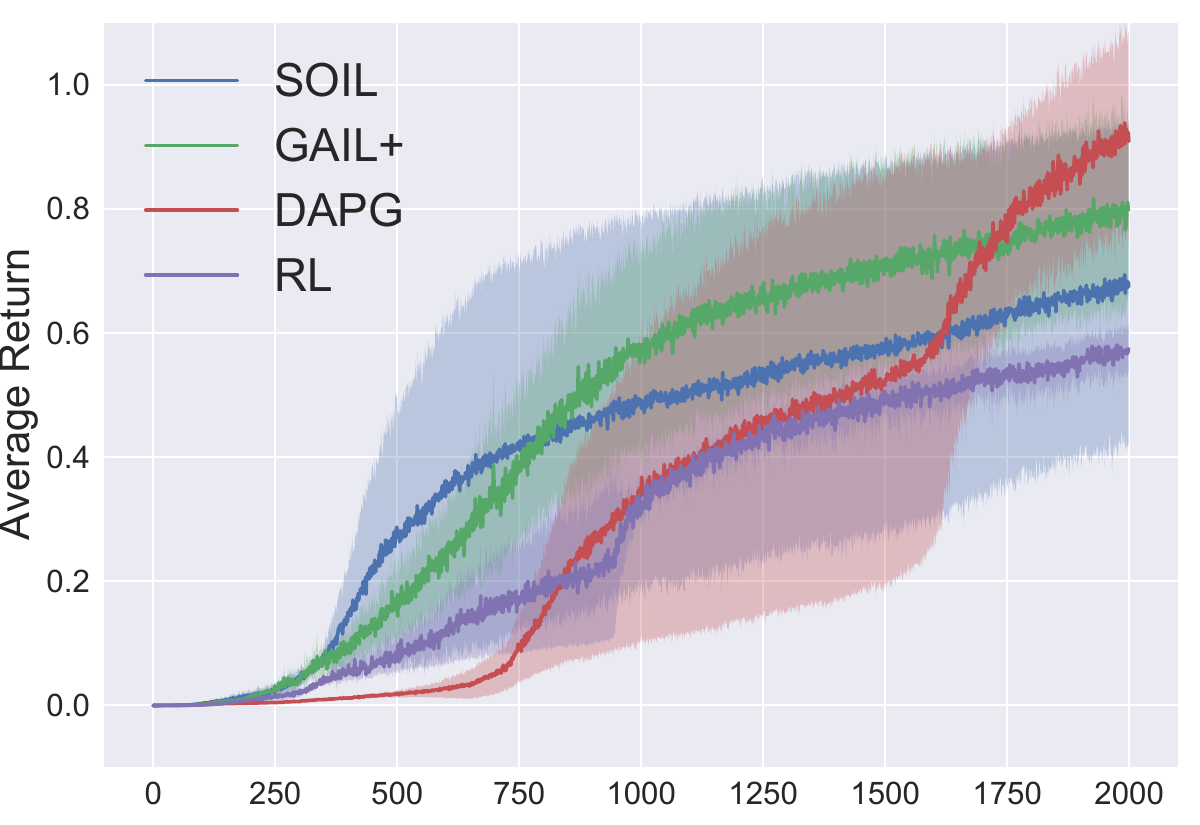}
		\caption*{\small{(a) Pour}}
	\end{minipage}
	\begin{minipage}{0.22\linewidth}
		\centering
		\vspace{0.6em}
		\resizebox{1\columnwidth}{!}{%
		\begin{tabular}{l|c}
         Model & Success (\%)\\\shline
         SOIL & $3.5\pm3.3$\\
         GAIL+ & $3.4\pm2.5$ \\
         DAPG & $\textbf{27.2}\pm\textbf{18.4}$ \\
         RL & $1.3\pm0.7$ \\
        \end{tabular}
        }
        \vspace{0.6em}
        \caption*{\small{(b) Pour}}
	\end{minipage}\hfill
	\begin{minipage}{0.27\linewidth}
		\centering
		\includegraphics[width=0.95\linewidth]{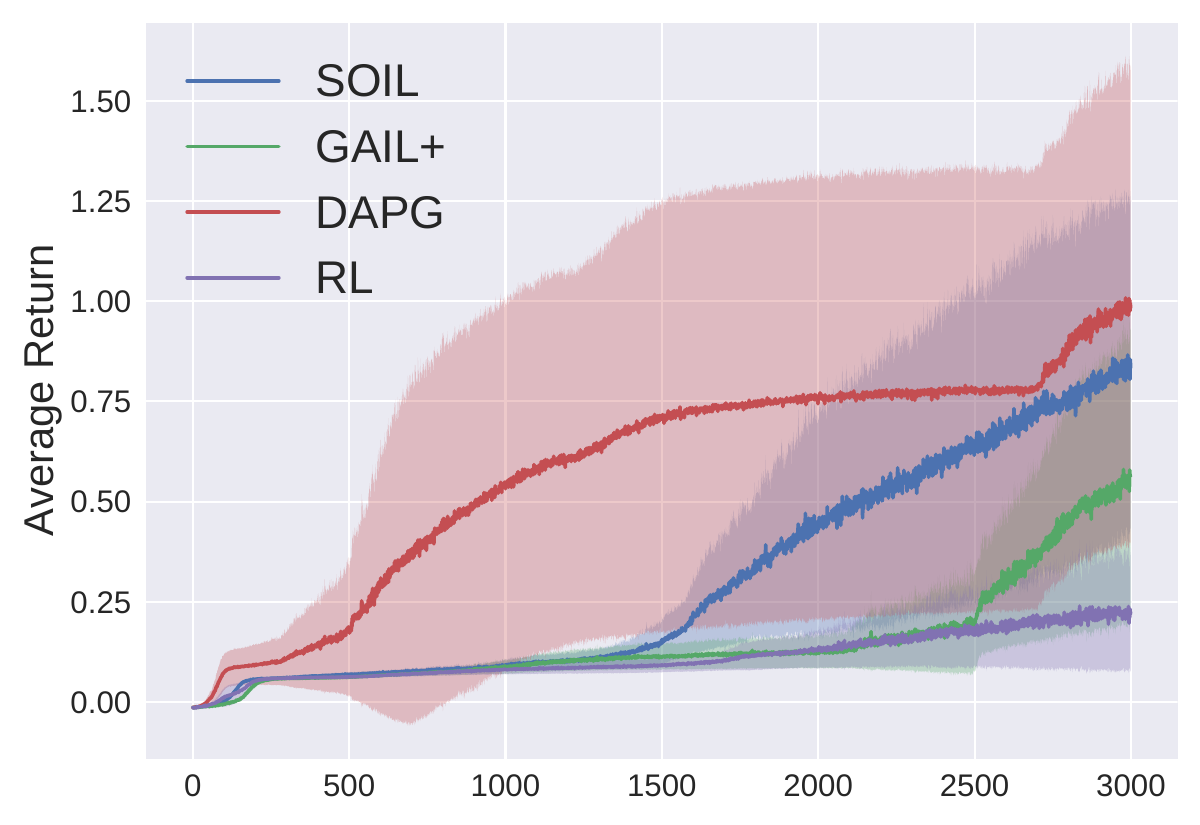}
		\caption*{\small{(c) Place Inside}}
	\end{minipage}
% 	\hspace{-2em}
	\begin{minipage}{0.22\linewidth}
		\centering
		\vspace{0.6em}
		\resizebox{1\columnwidth}{!}{%
		\begin{tabular}{l|c}
         Model & Inside Score \\\shline
         SOIL & $27.9\pm26.5$\\
         GAIL+ & $16.0\pm14.2$ \\
         DAPG & $\textbf{31.3}\pm\textbf{30.0}$ \\
         RL & $3.2\pm5.6$ \\
        \end{tabular}
        }
        \vspace{0.6em}
        \caption*{\small{(d) Place Inside}}
	\end{minipage}
	\vspace{-1em}
	\captionof{figure}{\small{\emph{Pour and Place Inside}. (a) learning curves of \emph{Pour}; (b) success rate of \emph{Pour}; (c) learning rate of \emph{Place Inside}; (d) Inside score of \emph{Place Inside}.}}
    \vspace{-1em}
    \label{fig:pour_place} 
\end{table}

\subsection{Experiments with Pour}
\label{sec:pour}
\vspace{-0.05in}

The \emph{Pour} task involves a sequence of dexterous manipulations: reaching the mug, holding and stably moving the mug, and pouring water into the container. We benchmark four imitation learning algorithms in Figure~\ref{fig:pour_place} (a) and (b). We observe that DAPG converges to a good policy with much fewer iterations: $27.2\%$ of the particles are poured into the container on average. Without the demonstrations, pure RL will only have a very small chance to pour a few particles into the container. State-action method DAPG performs better than state-only method SOIL. It is challenging to learn inverse model with water particles in this task for SOIL, while the analytical inverse dynamics function still provides reasonable actions for DAPG. See our videos and website for visualization. 

\subsection{Experiments with Place Inside}
\label{sec:place}
\vspace{-0.05in}
The \emph{Place Inside} task requires the robot hand to first pick up a banana, rotate it to the appropriate orientation, and place it inside the mug. We define a metric \emph{Inside Score}, which is computed based on the volume percentage of the banana inside of the mug.
We benchmark the imitation learning algorithms in Figure~\ref{fig:pour_place} (c) and (d). We find that DAPG outperforms other approaches whereas RL hardly learns to manipulate the object. Between state-only method (SOIL) and state-action method (DAPG), we again observe that computing action offline analytically achieves better results, due to the complexity of the task. 
\begin{table}[t]
    \begin{minipage}{0.33\linewidth}
		\centering
		\includegraphics[width=\linewidth]{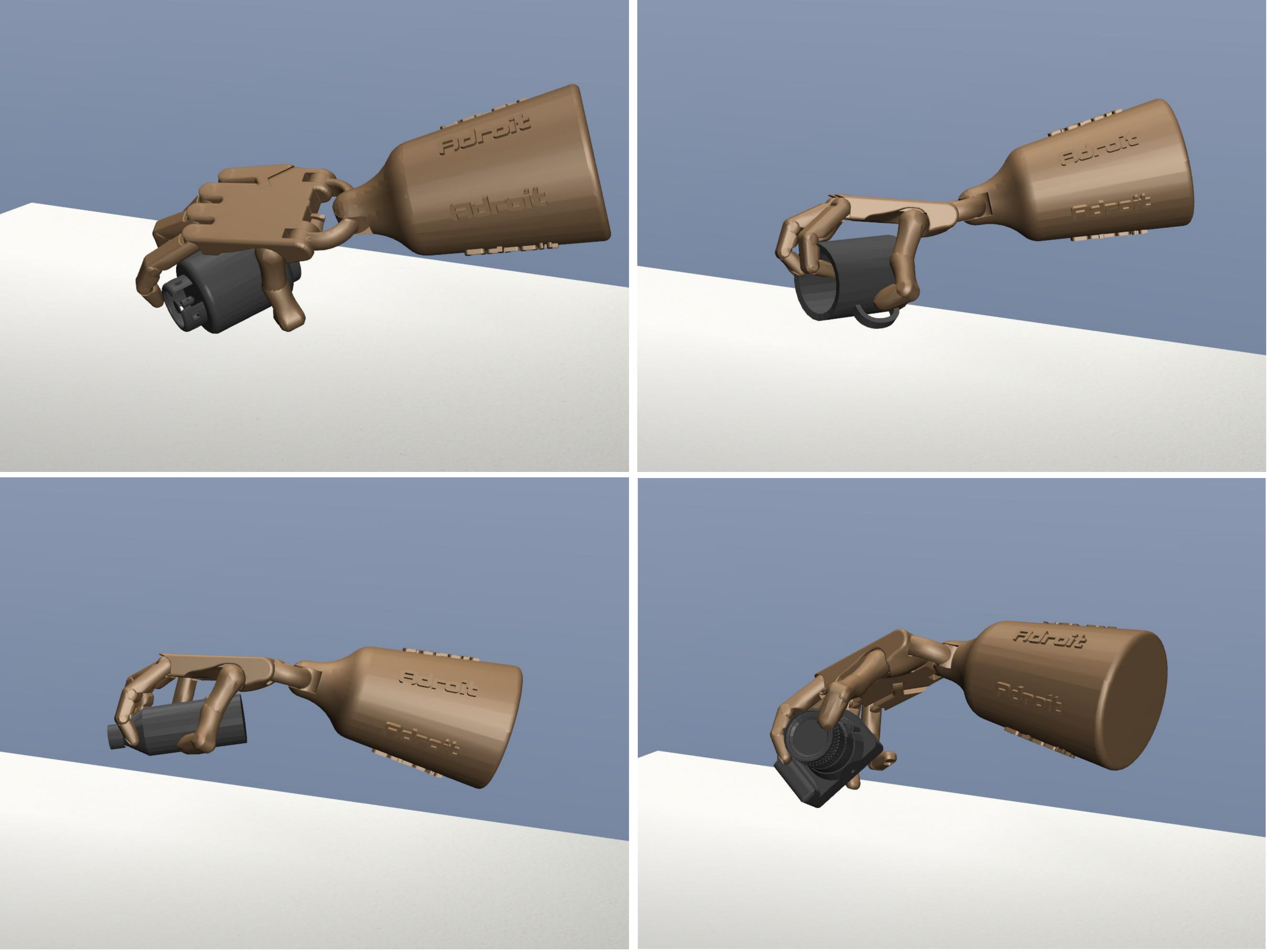}
	\end{minipage}\hfill
    \begin{minipage}{0.65\linewidth}
    \centering
    % \vspace{-0.5em}
    \resizebox{1\columnwidth}{!}{%
    \begin{tabular}{c|c|c|c|c} 
        \multicolumn{1}{c}{Setting} & \multicolumn{1}{c}{SOIL} & \multicolumn{1}{c}{}{GAIL+} & \multicolumn{1}{c}{DAPG} & \multicolumn{1}{c}{RL} \\ \shline
        \multicolumn{5}{c}{\textbf{Novel Object Instances, Same Category}} \\ \hline
        tomato.$\to$can & 62.0$\pm$27.3 & 30.7$\pm$23.5 & \textbf{83.6}$\pm$\textbf{4.7} & 15.9$\pm$16.7 \\ 
        mustard.$\to$bottle & 0.0$\pm$0.0& 0.0$\pm$0.0  & \textbf{68.5}$\pm$\textbf{7.0} & 0.0$\pm$0.0 \\ \
        mug.$\to$mug & 51.4$\pm$33.2 & 33.8$\pm$36.1 & \textbf{79.4}$\pm$\textbf{11.9} & 27.7$\pm$38.8 \\ 
        \multicolumn{5}{c}{} \\
        \multicolumn{5}{c}{\textbf{Novel Category}} \\ \hline
        tomato.$\to$cam. & 27.1$\pm$13.5 & 5.9$\pm$6.3 & \textbf{47.2}$\pm$\textbf{2.5} & 3.8$\pm$2.1 \\ 
        mustard.$\to$cam. & 2.6$\pm$2.8 & 0.0$\pm$0.0 & \textbf{49.9}$\pm$\textbf{8.5} & 0.1$\pm$0.1 \\ 
        mug.$\to$cam. & 25.8$\pm$18.8 & 17.3$\pm$23.1 & \textbf{33.4}$\pm$\textbf{3.9} & 7.5$\pm$9.5 \\
    \end{tabular}
    }
    \end{minipage} 
    \captionof{figure}{\small{\emph{Relocate} generalization. Left: visualization of \emph{Relocate} with unseen ShapeNet objects from 4 categories. Right: success rate of \emph{Relocate}. A$\to$B denotes using the policy trained on YCB object A to evaluate the performance on ShapeNet category B.}}
    \label{tab:generalization}
    \vspace{-2em}
\end{table}

\begin{table}[t]
\vspace{-0.1in}
	\begin{minipage}{0.44\linewidth}
		\centering
		\includegraphics[width=\linewidth]{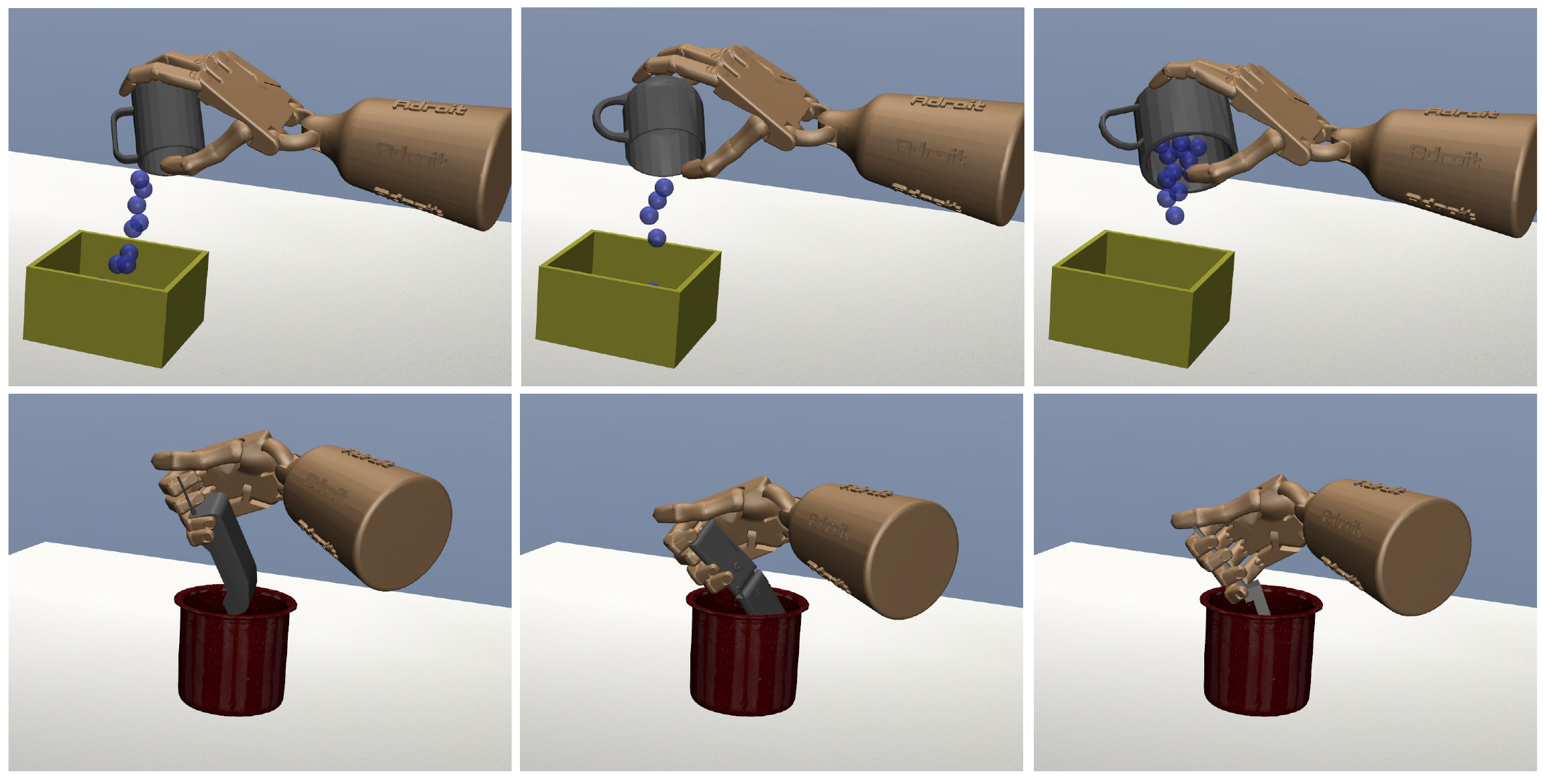}
		\caption*{\small{(a) Visualization of Two Tasks}}
	\end{minipage}\hfill
	\begin{minipage}{0.27\linewidth}
		\centering
		\vspace{1em}
		\resizebox{0.93\columnwidth}{!}{%
		\begin{tabular}{c|c}
         Model & Success  (\%) \\\shline
         SOIL & $3.8\pm3.2$\\
         GAIL+ & $2.1\pm0.8$ \\
         DAPG & $\textbf{23.3}\pm\textbf{10.4}$ \\
         RL & $0.5\pm0.3$ \\
        \end{tabular}}
        \vspace{1em}
        \caption*{(b) Pour}
        \label{table:student}
	\end{minipage}
	\begin{minipage}{0.27\linewidth}
		\centering
		\vspace{1em}
		\resizebox{1\columnwidth}{!}{%
		\begin{tabular}{c|c}
         Category & Inside Score \\\shline
         SOIL & $21.6\pm16.1$\\
         GAIL+ & $10.9\pm4.6$ \\
         DAPG & $\textbf{26.9}\pm\textbf{17.3}$ \\
         RL & $1.8\pm1.5$ \\
        \end{tabular}
        }
        \vspace{1em}
        \caption*{(c) Place Inside}
	\end{minipage}
    \captionof{figure}{\small{Generalization with \emph{Pour} and \emph{Place Inside}. (a) visualization of \emph{Pour} (top row) with ShapeNet mug, and \emph{Place Inside} with ShapeNet cellphone; (b) success rate of \emph{Pour} with ShapeNet Mug; (c) inside score of \emph{Place Inside} with ShapeNet cellphone.}}
    \label{fig:generalization_pour_place}
    \vspace{-3em}
\end{table}

\subsection{Generalization on Novel Objects and Category}

\textbf{Generalization with Relocate.}
We benchmark the generalization ability of the trained policy with different imitation learning algorithms. We directly deployed the trained policies for the \emph{Relocate} task using our demonstrations on: (i) novel unseen objects within the same category; (ii) object instances from a new category. For evaluation on unseen object instances within the same category, we choose the can, bottle, and mug categories from ShapeNet~\cite{chang2015shapenet} dataset, which corresponds to the tomato soup can, mustard bottle, and mug in our demonstrations. We omit evaluation on the clamp and sugar box policies since there are no corresponding categories in ShapeNet. For the evaluation on new category, we choose the camera category from ShapeNet. We select 100 object instances for each category and pre-process the objects to fit the size of robot hand. More details about data pre-processing can be found in supplementary. We report the results in Figure~\ref{tab:generalization}. The success rate is computed as a mean accuracy over all object instances within a category. We find DAPG generalizes well on unseen objects in both same category and novel category experiments, significantly outperforming other approaches. We can also observe generalization to novel categories brings a larger challenge. 

\textbf{Generalization with Pour.} We benchmark the generalization performance of the trained policies from all four methods on new mug instances for \emph{Pour} task. We choose 100 object instances from the ShapeNet mug category for deployment. The top row of Figure~\ref{fig:generalization_pour_place} (a) visualizes the task with different mugs and Figure~\ref{fig:generalization_pour_place} (b) shows the success rate. In this benchmark, DAPG still performs the best across all methods. Interestingly, the performance is not much worse than \emph{Pour} with the trained object (by comparing Fig.~\ref{fig:pour_place} (b)). This shows the robustness of imitation learning using our pipeline. 

\textbf{Generalization with Place Inside.}
For \emph{Place Inside}, we replace the YCB banana using the objects from cellphone category in ShapeNet as visualized in the bottom row of Figure~\ref{fig:generalization_pour_place} (a). Figure~\ref{fig:generalization_pour_place} (c) shows the generalization performance when directly deploying the policy trained with banana. We find the policy performance is also close to Place Inside with banana (by comparing Fig.~\ref{fig:pour_place} (d)). This indicates the robustness of policy against small shape variations and human demonstration on single object can also benefit the tasks on new objects. 

%% file: sections/conclusion.tex
\vspace{-0.05in}
\section{Conclusion}
\vspace{-0.05in}

To the best of our knowledge, DexMV is the first work to provide a platform on computer vision/simulation systems and a pipeline on learning dexterous manipulation from human videos. We propose a novel demonstration translation module to bridge the gap between these two systems. We hope DexMV opens new research opportunities for benchmarking imitation learning algorithms for dexterous manipulation.

%% file: sections/supp_section.tex
\section{Appendix Overview}
This supplementary material provides more details, results and visualizations accompanying the main paper. 
In summary, we include
\begin{itemize}
    \item More details about video data collection;
    \item More details about the \emph{Relocate}, \emph{Pour}, and \emph{Place inside} environments
    \item More details about demonstration translation;
    \item More visualization of hand-object pose estimation and hand motion retargeting results.
\end{itemize}

\begin{figure*}[t!]
    \centering
    \includegraphics[width=\linewidth]{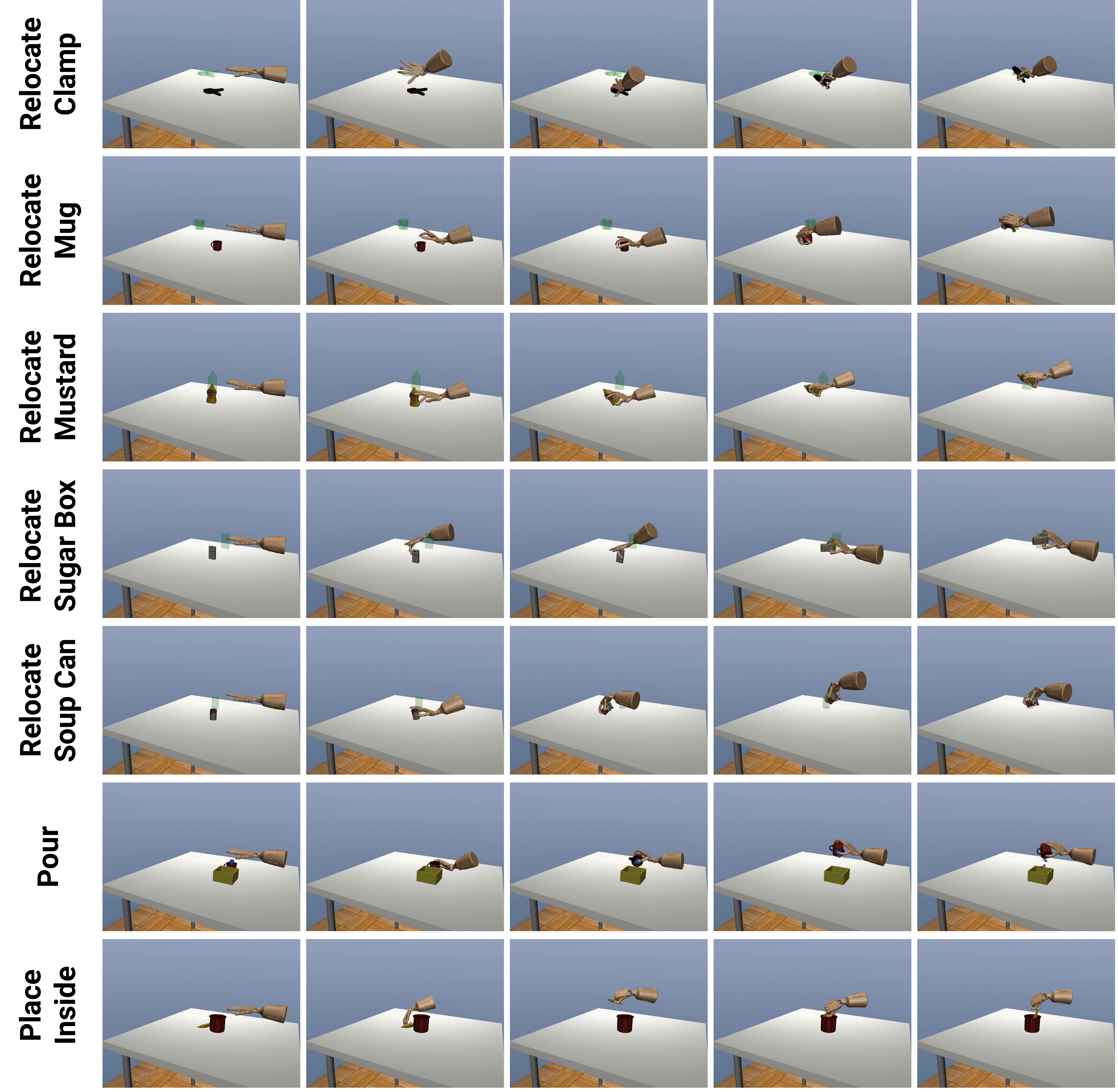}
    \caption{\titlecap{DexMV Tasks.}{There are three types of tasks in the figure. The first five columns: \emph{relocate} with mug, mustard bottle, clamp, sugar box, tomato soup can. \emph{Relocate} task means that the agent needs to move the object from the initial position to the target position. The sixth and seventh columns: \emph{Pour} and \emph{Place Inside}. In \emph{Pour} task, the agent needs to pour the water particles inside a mug into the yellow container. In \emph{Place Inside} task, the agent needs to manipulate the orientation of the banana to place it inside a mug. Both of the last two tasks require delicate manipulation.}}
    \label{fig:supp_teaser}
    \vspace{-2mm}
\end{figure*}

\section{Video Data Collection}
We use Intel RealSense D435 cameras to collect human demonstrations of manipulating different objects for finishing diverse tasks on the table. In detail, each captured demonstration is about $10$ seconds. Moreover, after the demonstration is captured, only the pour task is in need to reset the particles back into the mug, which tasks about $6$ seconds. Thus, the time cost for data collection is not high, and the procedure tends to be scalable on different objects and tasks. In practice, it takes about $60$ minutes to capture all $100$ sequences for a task.

\section{Environments}
\label{sec:supp_env}
We propose three types of manipulation tasks along with the DexMV Platform: \emph{Relocate}, \emph{Pour}, and \emph{Place Inside}. The manipulated objects come from YCB Dataset~\cite{calli2015benchmarking}. Environments use the MuJoCo simulator~\cite{todorov2012mujoco} with timestep set to $0.002$ and frame skip set to $5$. We adopt the same contact friction parameters following the setting in the literature~\cite{rajeswaran2017learning}. We use the open-source MuJoCo model of Adroit Hand~\footnote{https://github.com/vikashplus/Adroit}. Figure~\ref{fig:supp_teaser} shows the seven different task used in DexMV:\emph{Relocate} with five different objects, \emph{Pour}, and \emph{Place Inside}.

\mypara{Action.} The action space is the same for all tasks, which is the motor command of $30$ actuators on the robotic hand. The first $6$ motors control the global position and orientation of the robot while the last $24$ motors control the fingers of the hand. We normalize the action range to $(-1, 1)$ based on actuator specification.

\subsection{Relocate}
\mypara{Observation.} The observation of \emph{Relocate} is composed of four components: (i) joint angles of adroit robotic hand; (ii) global position of adroit hands root; (iii) object position; (iv) target position. The overall observation space is $39$-dim.

\mypara{Reward.} The reward is defined based on three distances: (i) the distance between the robot hand and the object; (ii) the distance between the robot hand and the target; (iii) the distance between the object and the target. Lower distance corresponds to higher reward.

\mypara{Reset.} For each episode, the xy position of both object and target is randomized within a $(-0.3, 0.3)$ square on the table. The height of target is randomized between $(0.15, 0.25)$.

\subsection{Pour}\label{subsection:pour}
\mypara{Observation.}
Similar to \emph{Relocate}, we include the robot joint angles, root position of robot hand, and object position in the observation. In \emph{Pour}, we replace the target position with the container position. Besides, since the agent needs orientation information of the mug to pour the water particles, we add a quaternion to represent object orientation.

\mypara{Reward.} The main reward is based on the final number of particles that fell within the container. Additionally, similar to \emph{Relocate}, we use the distance between the robot hand and the object as well as the distance between the object and the container to provide part of the rewards. Lower distance leads to higher reward. The coefficient of the main reward is $10\times$ larger than the reward computed based on the distance. 

\mypara{Reset.} For each episode, the xy position of the mug is randomized between $(-0.1, 0.1)$ on the table. The water particles are inside the mug at the beginning of each episode. The container is always at the center of the tabletop.

\subsection{Place Inside}\label{subsection:place}
\mypara{Observation.}
The observation space of the \emph{Place Inside} task is the same as the observation space of \emph{Pour} as described in Section~\ref{subsection:pour}.

\mypara{Reward.} The main reward is based on the position and orientation of the manipulated object. If the object is placed inside of the mug, the agent will get a large portion of the reward. Similar to \emph{Relocate}, we also add a lifting reward to encourage the robot to first lift the object before moving it towards the container. 

\mypara{Reset.} For each episode, the xy position of the object is randomized between $(-0.15, 0.15)$  on the table. The container, i.e. mug, is always placed at the center of tabletop.

\subsection{ShapeNet Objects}
In the \emph{Generalization on Novel Objects and Category} experiments (Section 8.4) of the main paper, we use objects from ShapeNet~\cite{chang2015shapenet} dataset. We pre-process the ShapeNet geometry by scaling the mesh so that it can be used in our simulated environment. The YCB objects~\cite{calli2015benchmarking} are captured from the real scan, so the scale of object mesh from YCB dataset aligns with the real counterpart and can be used for robot manipulation directly. Different from the YCB dataset, the ShapeNet dataset does not contain any scale information, e.g. the mug in ShapeNet can be larger than the robot. So we need to scale the object to a reasonable size in which it can be manipulated by the robot. We scale the object based on the diagonal length of the object bounding box. For each category, we manually select a diagonal length so that all object instances from the category will have the same bounding-box diagonal length after scaling. Besides, we will not use objects with non-manifold geometry to avoid instability in physical simulation. For the \emph{Place Inside} task, the object mesh should be watertight for volume computation. We use the convex meshes processed by VHACD~\cite{mamou2009simple} for both simulation and volume computation.

\section{Demonstration Translation}
\label{sec:supp_demo}
\subsection{Kinematics Model}
In Table \ref{tab:supp_kinematics}, we compare the difference of kinematics model between the human hand and robot hand. The overall Degree-of-Freedom(DoF) of the human hand is higher than the DoF of the robot hand. Thus hand motion retargeting from human to robot is projecting a pose from a higher dimension to a lower dimension, which will lose information inevitably. The motion retargeting module try to maintain the task space vectors between these two different kinematics model. 

\begin{table}[]
    \centering
    \begin{tabular}{c|c|c|c|c}
         & \textbf{H}-Joints & \textbf{H}-DoF & \textbf{R}-Joints & \textbf{R}-DoF \\
          \shline
        Thumb & 3x Ball & 9 & 5x Revolute & 5 \\
        Index & 3x Ball & 9 & 4x Revolute & 4 \\
        Middle  & 3x Ball & 9 & 4x Revolute & 4 \\
        Ring  & 3x Ball & 9 & 4x Revolute & 4 \\
        Pinkie & 3x Ball & 9 & 5x Revolute & 5 \\
        Wrist & Null & 0 & 2x Revolute & 2 \\
        Root & 1x Free & 6 & 1x Free & 6 \\
        Overall & N/A & 51 & N/A & 30
    \end{tabular}
    \vspace{0.3em}
    \caption{\titlecap{Comparison of Kinematics}{We compare the kinematic model between MANO human hand and the robot hand we used in simulator. \textbf{H} is the abbr for Human while \textbf{R} stands for robot. For example, the \textbf{H}-Joints column shows the number and type of joints for a specific sub-part in the kinematics model. Each ball joint has 3 Degree-of-Freedom(DoF), each revolute joint has 1 DoF, and each free joint has 6 DoF}}
    \vspace{-0.2in}
    \label{tab:supp_kinematics}
\end{table}

\subsection{Initialization in Hand Motion Retargeting}

As mentioned in the Demonstration Translation section of the main paper, the robot joint angles are solved using optimization. A good initialization is essential for optimization. For $t \ge 1$, $q_t$ is initialized using the optimization results $q_{t-1}$. Here we will discuss how to initialize $q_t$ for $t=0$. Previous work~\cite{handa2020dexpilot} initializes $q_t$ with zero vectors. The optimization cannot provide reasonable outputs when the goal is far from zeros. To tackle this issue, we use a heuristic function $\phi(\theta_0)$ to initialize $q_0$. The heuristic function takes the MANO hand pose parameters $\theta_0$ and outputs an estimation of the robot joint angles. The heuristic function $\phi(\theta_0)$ can be regarded as a coarse hand motion retargeting function which does not consider the shape difference between human and robot hands. It is only used to provide a reasonable initialization for further optimization.

Given the axis-angle representation $\theta_0$ from hand pose estimation, we first convert it to $15$ rotation matrices. Then we compute the projection of each rotation matrix on the joint axis direction of each robot hand joint. Then we use a manually-designed vector $w$ to map the projection to the robot joint angle. The overall function can be formulated as below,

\begin{equation}
    \phi(\theta_0) = \mathbf{P_{rot}}(\mathbf{R}(\theta_0)) w
\end{equation}

where $\mathbf{R}(\cdot)$ is the Rodrigues' formula that maps axis-angle to rotation matrix, $\mathbf{P_{rot}}(\cdot)$ is the projection function to compute the nearest rotation along with the robot joint direction for each joint, i.e. projection. $\mathbf{P_{rot}}(\mathbf{R}(\theta_0)) \in \mathbb{R}^{24 \times 15}$ and $w \in \mathbb{R}^{15}$ is a manually designed vector. Thus the dimension of $\phi(\theta_0)$ is $24$, which corresponds to the joint angles for finger and wrist. As mentioned in Table~\ref{tab:supp_kinematics}, the overall DoF of the robot hand is $24 + 6 = 30$, which includes $6$ DoF hand root pose. We directly use the root position plus root orientation from human hand pose estimation as the root pose for the robot hand.

\subsection{Post Processing}
\mypara{Filter Estimated Pose:} When human is manipulating the object, either hand or object is in heavy occlusion, which may cause inconsistent estimation results. To improve the temporal consistency of the estimated hand and object poses, we apply a digital low-pass filter to remove the high-frequency noise. The sampling frequency of the filter is $100$ while the cutoff frequency is $5$ for the position of both object and hand. Filtering the rotation is not as straightforward as filtering the position. To get a smooth orientation sequence, we first convert the rotation into $so(3)$ lie algebra. Then, we apply the filter in $so(3)$ space and convert it back to rotation matrix $SO(3)$ after filtering.

\mypara{Hindsight Goal Position for Relocate.} 
As mentioned in the Task Section of the main paper, \emph{relocate} is a goal-conditioned task. The goal information should also be included in the state representation. To provide goal information from human demonstration, we use the position of the object in the last step as the hindsight goal.

\mypara{Frame Alignment.}
Spatial quantities like object pose are dependent on the frame in which it is observed. The natural frame for pose estimation results is the camera frame. In the simulated environment, such a camera frame does not exist and the pose is represented in the world coordinate fixed on the table. Thus we also align the frame in demonstrations to match the simulated environment.

\begin{figure*}[t!]
    \centering
    \includegraphics[width=\linewidth]{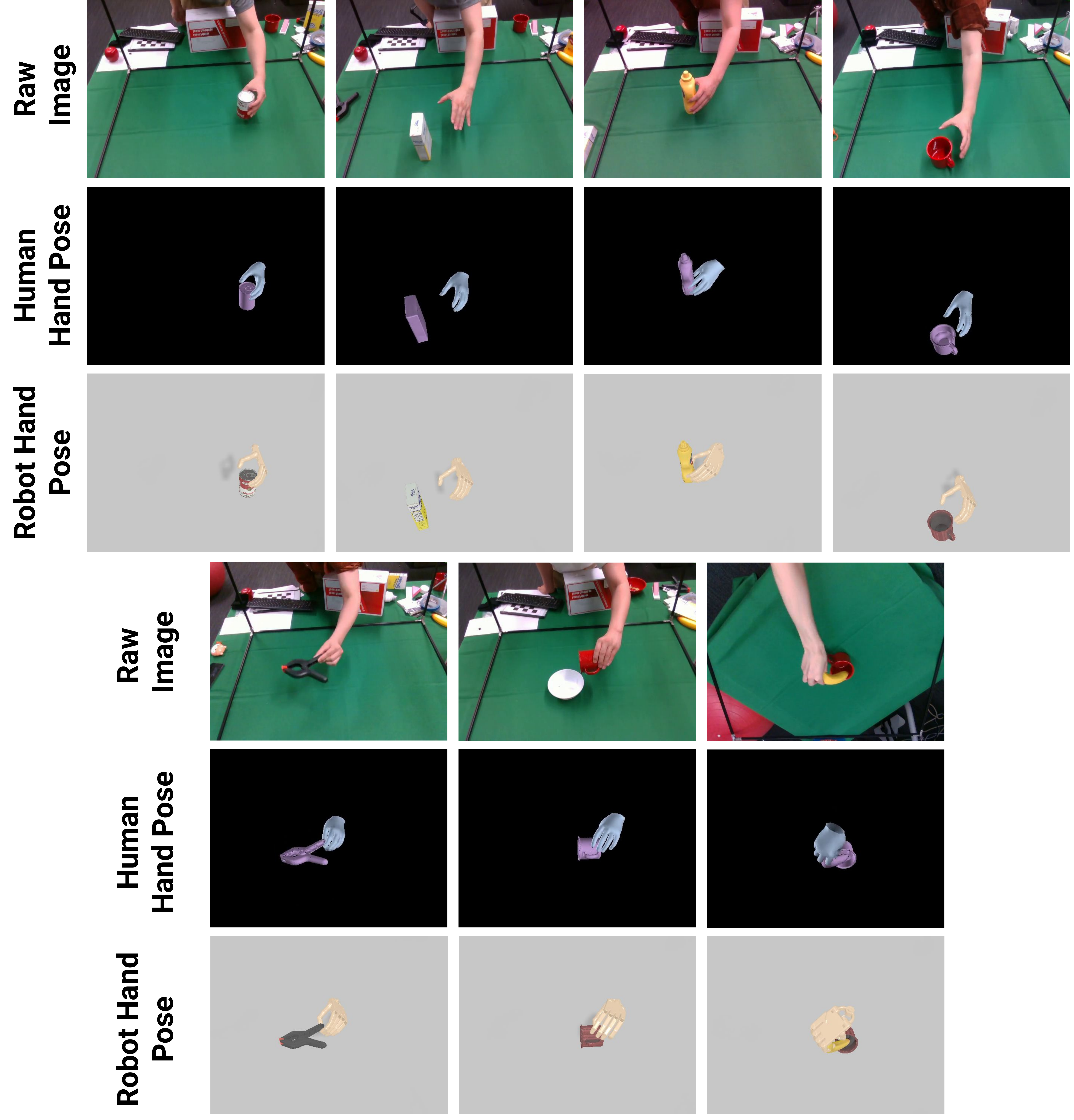}
    \caption{\titlecap{3D hand-object pose estimation results and hand motion retargeting results.} Visualization on relocate tomato soup can, sugar box, mustard bottle, mug, clamp, pour, and place inside.}
    \label{fig:sup_viz}
    \vspace{-2mm}
\end{figure*}
\section{More Visualization of Hand-Object Pose Estimation and Hand Motion Retargeting}
In this section, we provide more visualization on hand-object pose estimation and hand motion retargeting in Figure~\ref{fig:sup_viz}. The four tasks in the first three rows are \emph{Relocate} with tomato soup can, sugar box, a mustard bottle, and mug. The three tasks in the last three rows are: \emph{Relocate} with clamp, \emph{Pour}, and \emph{Place Inside}.

% {\small
% \bibliography{egbib}
% }